\pdfoutput=1 
\documentclass{article}

\usepackage[T1]{fontenc}
\usepackage[final]{corl_2026}
\usepackage{graphicx}
\usepackage{array}
\usepackage{booktabs}
\usepackage{multirow}
\usepackage{capt-of}

\makeatletter
\renewcommand{\@noticestring}{%
  4th Workshop on Learning Effective Abstractions for Planning (LEAP) at the
  10th Conference on Robot Learning (CoRL 2026), Austin TX, USA.%
}
\makeatother

\hypersetup{
  pdftitle={Toward Planning Representations That Separate Task Structure from Visual Content},
  pdfauthor={Aarav Srivastava, Vishwas Sathish, Rajesh P. N. Rao}
}

\title{Toward Planning Representations That Separate Task Structure from Visual Content}

\author{
  Aarav Srivastava, Vishwas Sathish, Rajesh P.\,N. Rao\\
  Paul G. Allen School of Computer Science \& Engineering\\
  University of Washington, Seattle, United States\\
  \texttt{\{aaravs07, vsathish, rao\}@cs.washington.edu}
}

\begin{document}
\maketitle


\begin{abstract}
Effective planning requires representations that preserve task-relevant
geometry and object identity while ignoring visual variation that does not
alter the task. Existing evaluations of visual generalization often change
several factors simultaneously, making the source of a failure difficult to
identify. We introduce a controlled evaluation of object-goal navigation
(ObjectNav), the task of reaching an instance of a named object category, in
procedurally generated ProcTHOR houses. Training and evaluation houses share
floor plans, object locations, goals, and start states, while visual factors
are changed individually or in combination. Across model-free reinforcement
learning (RL), model-based RL, and agents using frozen pretrained visual
representations, our results show that appearance changes impair navigation,
but their effects differ sharply by factor. Surface-material changes account
for most of the average change in success, whereas clutter and lighting have
little effect. The magnitude of a global image-statistics shift does not
reliably predict performance: changing only the goal object's appearance can
substantially impair an end-to-end agent despite a small image-level change.
We found that a model-based agent that plans over frozen features from the
pretrained DINOv2 image encoder achieves the highest success (0.83) under the
full shift, but remains sensitive to surface appearance. Our experiments reveal
an important requirement for planning representations: they must separate
invariant spatial task structure from appearance.
\end{abstract}



\section{Introduction}
\label{sec:intro}

A representation used for planning should assign similar states to scenes that
afford the same actions and outcomes. In object-goal navigation
(ObjectNav)~\cite{batra2020objectnav}, an agent must find and reach an instance
of a named object category, such as a bed. An agent trained to reach the bed in
one house should still reach it after a room is repainted if the navigable
space, object positions, goal, and start state are unchanged. Failure under
that change shows that the agent's behavior depends on visual information that
the navigation problem does not require.

Visual distribution shifts can degrade reinforcement learning (RL) policies
trained from pixels~\cite{cobbe2020procgen, kirk2023survey,
stone2021distracting, hansen2021soda, yuan2023rlvigen, laskin2020rad,
yarats2021drq, yarats2022drqv2}. Existing benchmarks commonly change several
properties together, including layout, objects, lighting, camera conditions,
and appearance. They establish whether an agent generalizes, but usually cannot
determine which changed property caused a failure or whether separate failures
combine. For planning, this distinction matters. A latent state may support
planning in the training environment while still binding an action-relevant
object or a route to the particular textures, meshes, or illumination seen
during training.

\begin{figure}[t]
  \centering
  \includegraphics{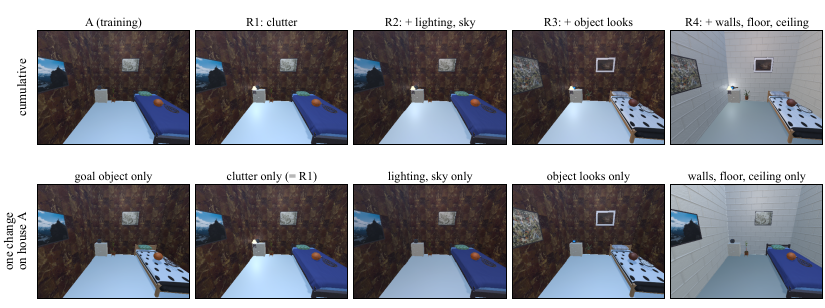}
  \caption{Controlled visual changes in one ProcTHOR house. (Top row) The
  training house A and the four cumulative rungs. (Bottom row) Variants that
  apply only one change to house A while preserving the floor plan, object
  positions, goal location, and evaluation start poses.}
  \label{fig:views}
\end{figure}

What are the properties that underlie current failures in visual generalization
in planning and RL? We address this question with a controlled evaluation of RL
navigation agents in ProcTHOR~\cite{deitke2022procthor}, a collection of
procedurally generated houses from the AI2-THOR
simulator~\cite{kolve2017ai2thor}. For each training house, the evaluation
houses preserve the reachable floor area, object positions, goal instances, and
evaluation start poses. They differ in appearance, and an object swap also
changes that object's mesh; Figure~\ref{fig:views} shows one house and its
variants. We independently vary surface materials, lighting, sky, non-goal
object appearance, goal-object appearance, and clutter, then evaluate both
isolated changes and cumulative combinations. The setup turns visual
generalization into a diagnosis problem: it identifies which visual variables
each agent's behavior depends on.

We evaluate model-free agents trained with proximal policy optimization
(PPO)~\cite{schulman2017ppo}, PPO with visual augmentation, PPO policies on
frozen pretrained encoders, two model-based agents that learn a world model,
DreamerV3~\cite{hafner2025dreamerv3} and TD-MPC2~\cite{hansen2024tdmpc2}, and
TD-MPC2 with frozen DINOv2 features~\cite{oquab2024dinov2}. The results
identify two types of sensitivity. Across agents, changes to walls, floors, and
ceilings produce the largest average change in success. End-to-end PPO also
depends strongly on the goal object's particular appearance, even though
changing that object affects relatively few pixels. Global image-shift
magnitude therefore only partly characterizes the difficulty of a perturbation.
Frozen features from pretrained DINOv2 reduce sensitivity to lighting, sky, and
object appearance. TD-MPC2 trained on these features reaches the highest
success, although surface-material changes still reduce it. Together, these
results highlight the importance of separating invariant spatial task structure
from appearance for visual generalization in embodied planning.

We contribute (i) a controlled ObjectNav evaluation whose changed houses are
verified in the simulator to keep the floor area, goal, and start poses of the
training house; (ii) a comparison of model-free, model-based, and
frozen-encoder agents on it, including one frozen feature under both kinds of
learner; and (iii) the findings above. We propose no new
representation-learning method.


\section{Related Work}
\label{sec:related}

\paragraph{Visual generalization in RL.}
Procedural generation evaluates generalization to unseen
levels~\cite{cobbe2020procgen}; \citet{kirk2023survey} survey
zero-shot generalization in deep RL. The Distracting Control Suite, SODA, and
RL-ViGen vary visual factors such as backgrounds, color, lighting, camera
configuration, and layout~\cite{stone2021distracting, hansen2021soda,
yuan2023rlvigen}. These benchmarks measure performance under a changed test
distribution. Our evaluation instead keeps the navigation problem fixed and
isolates the contribution of each appearance factor. Controlled perturbation
benchmarks exist in imitation-learning manipulation: Factor
World~\cite{xie2024decomposing}, The Colosseum~\cite{pumacay2024colosseum}, and
KitchenShift~\cite{xing2021kitchenshift} vary factors such as lighting,
textures, and camera pose, and \citet{gao2024compositional} study composition
of observed factor values. We apply this design to RL navigation, where no
value of a changed factor is seen in training, compare model-free,
model-based, and frozen-encoder agents, and test whether separately measured
effects combine.

\paragraph{Augmentation and pretrained representations.}
Domain randomization and image augmentation expose policies to selected
nuisance variation during training~\cite{tobin2017domain, laskin2020rad,
yarats2021drq, yarats2022drqv2}. Frozen visual encoders pretrained on large
image or video corpora are also used in imitation learning and model-free
RL~\cite{parisi2022unsurprising, nair2022r3m, majumdar2023vc1}, although
well-tuned learning-from-scratch baselines can be
competitive~\cite{hansen2023pretraining}. \citet{burns2024pvr}
relate performance under visual shift in manipulation to segmentation
properties of visual encoders. Results for frozen features in model-based RL
differ across settings: \citet{schneider2024pvrmbrl} report
limited benefits, while \citet{jones2025pvm} report benefits under
severe visual shift. We place one frozen DINOv2 feature under a model-free and
a model-based learner.

\paragraph{Abstractions for planning.}
Task-informed objectives can direct a latent state to retain information needed
for control while discarding nuisance detail~\cite{fu2021tia, zhang2021dbc}.
DreamerV3 reconstructs observations, whereas TD-MPC2 learns latent dynamics,
reward, and value without an observation decoder. Comparing them under
controlled shifts tests whether these modeling choices are sufficient to remove
appearance dependence. Our result does not isolate a single training objective
as its cause. It identifies the visual variables that an abstraction for
planning must handle.


\section{Controlled ObjectNav Evaluation}
\label{sec:eval}

This section defines the task, the appearance interventions, and the
measurements.

\subsection{Task and invariants}
\label{sec:task}

We use object-goal navigation~\cite{batra2020objectnav} in
AI2-THOR~\cite{kolve2017ai2thor} with ProcTHOR-10k
houses~\cite{deitke2022procthor}. The agent observes a $128 \times 128$ RGB
image with a $90^\circ$ field of view and selects one of five actions: move
forward by 0.25\,m, turn left or right by $30^\circ$, or tilt the camera up or
down by $30^\circ$. An episode succeeds when the goal object is visible and
within 1.5\,m in straight-line distance. Episodes end after 200 steps. The reward
is $+10$ on success, $-0.01$ per step, and $+1$ per meter of reduction in
straight-line distance to the goal.

We select five houses, P0--P4, with 128--681 reachable floor positions on a
0.25\,m grid. Each house has a fixed goal-object type and location. Twenty
percent of reachable positions are held out from training. We draw 25
position-and-heading start poses once from the held-out set and use exactly
those poses in every evaluation house. This controls the task distribution as
well as the geometry.

\subsection{Appearance interventions}
\label{sec:interventions}

Starting from a training house A, we construct variants that change one of six
factors: surface materials for walls, floors, and ceilings; lighting; sky;
non-goal object appearance; goal-object appearance; or clutter. Object changes
replace an object with a different asset of the same type. The goal asset is
changed only when a replacement with the same footprint exists. Clutter
consists of two to eight small objects placed on existing surfaces.

We use two complementary evaluations. The reordered cumulative ladder starts
with clutter (R1), adds lighting and sky (R2), adds object appearance including
the goal (R3), and then adds surface materials (R4). The order follows the
average effect measured in single-factor trials, from smallest to largest. The
original ladder begins with surfaces, lighting, and sky (L1), adds object
appearance (L2), then adds clutter (L3). Each ladder combines the same
per-factor fields. A factor therefore has the same implementation wherever it
appears, and R4 and L3 are the same house.

Each constructed house must satisfy three checks in the simulator: it has the
same reachable positions as A, each evaluation start pose remains reachable,
and it contains the same number of goal instances. We run all three checks on
three independent simulator loads (one load for the goal-control houses of
Appendix~\ref{app:goal}) as a guard against simulator nondeterminism. All
objects are kinematic, so the agent cannot move them. Replacing a goal asset
can change the shortest distance to its surface by up to 0.49\,m, so
path-efficiency measures such as success weighted by path length
(SPL)~\cite{anderson2018evaluation} use distances measured once in A; we report
success rates throughout. Surface, lighting, sky, and clutter changes leave
path lengths to the goal unchanged; only object swaps alter them
(Appendix~\ref{app:houses}).

\begin{figure}[t]
  \centering
  \includegraphics{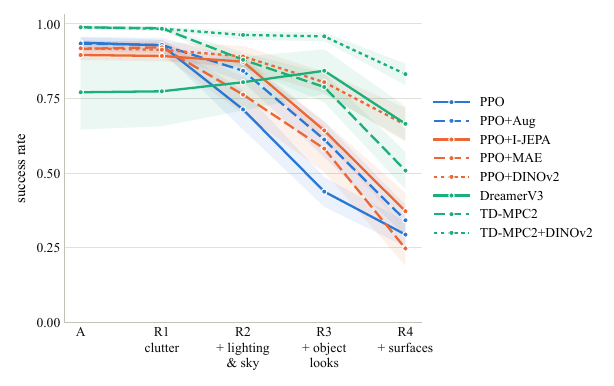}
  \caption{Success rate along the reordered ladder, 25 runs per agent type.
  Rungs are cumulative: R2 is clutter plus lighting and sky. Bands: 95\%
  confidence intervals. Values: Appendix~\ref{app:ladder}.}
  \label{fig:reordered}
\end{figure}

\subsection{Evaluation and image measurements}
\label{sec:measurements}

Agents are evaluated without further training, from the same start poses before
and after each intervention. Tables give success as the mean $\pm$ standard
deviation over the five training seeds, each seed averaged over houses; figures
give 95\% confidence intervals from a stratified bootstrap that resamples the
seeds within each house~\cite{agarwal2021precipice}. Declared significance tests
between agent types appear in Appendix~\ref{app:declared}. Five DreamerV3 runs
did not learn the training house (success below 0.5); analyses that compare a
change with house A exclude them, while raw success rates keep every run.

For each changed house, we pool pixels from the 25 start views into one 256-bin
histogram per color channel and measure its Wasserstein-1 (earth mover's)
distance from A in gray levels (0--255). We also measure the mean per-pixel
difference between aligned views. Neither depends on the agent.


\section{Agents and Training}
\label{sec:agents}

All agents train for 300k environment steps in house A, with five seeds in each
of five houses. PPO~\cite{schulman2017ppo} uses the convolutional encoder of
\citet{mnih2015dqn}, trained end to end through
Stable-Baselines3~\cite{raffin2021sb3}. PPO+Aug applies brightness, contrast,
saturation, and hue jitter during training. PPO+I-JEPA, PPO+MAE, and PPO+DINOv2
train only a small policy and value head on a frozen vision transformer (ViT)
pretrained by I-JEPA~\cite{assran2023ijepa}, MAE~\cite{he2022mae}, or
DINOv2~\cite{oquab2024dinov2}.

DreamerV3 learns a recurrent latent world model and reconstructs $64 \times 64$
observations~\cite{hafner2025dreamerv3}. TD-MPC2 learns latent dynamics,
reward, and value without an observation decoder, then plans with model
predictive path integral (MPPI) control from a three-frame
stack~\cite{hansen2024tdmpc2}. TD-MPC2+DINOv2 instead receives the frozen
DINOv2 feature of one frame, as PPO+DINOv2 does, through its multilayer
perceptron (MLP) state encoder. Appendix~\ref{app:settings} gives all settings.
The agents span four choices relevant to abstraction learning: end-to-end
versus frozen vision, model-free control versus latent planning, predictive
versus reconstructive latent modeling, and augmentation during training.


\section{Results}
\label{sec:results}

Each effect of a change below compares an agent with itself in house A and in
the changed house (same weights, same start poses), so differences between
agent types in input, encoder, or capacity cannot create it; they matter only
when agent types are compared, mainly in Section~\ref{sec:pretrained}.

\subsection{Accumulated appearance changes reduce navigation success}
\label{sec:accumulated}

\begin{figure}[t]
  \begin{minipage}[t]{0.31\textwidth}
    \vspace{0pt}
    \captionof{table}{\raggedright Success after one factor changes: mean over the eight
    agent types (runs that learned house A) $\pm$ standard deviation over
    seeds. Bold: highest and lowest. ``Goal object'': the four houses with a
    swappable goal.}
    \label{tab:single}
    \centering
    \small
    \begin{tabular}{@{}lc@{}}
      \toprule
      Change & Success \\
      \midrule
      None (house A) & 0.94\,$\pm$\,0.02 \\
      \midrule
      Clutter       & \textbf{0.94}\,$\pm$\,0.01 \\
      Lighting      & 0.92\,$\pm$\,0.01 \\
      Sky           & 0.89\,$\pm$\,0.01 \\
      Other objects & 0.90\,$\pm$\,0.02 \\
      Goal object   & 0.86\,$\pm$\,0.04 \\
      Surfaces      & \textbf{0.66}\,$\pm$\,0.02 \\
      \bottomrule
    \end{tabular}
  \end{minipage}\hfill
  \begin{minipage}[t]{0.66\textwidth}
    \vspace{0pt}
    \centering
    \includegraphics{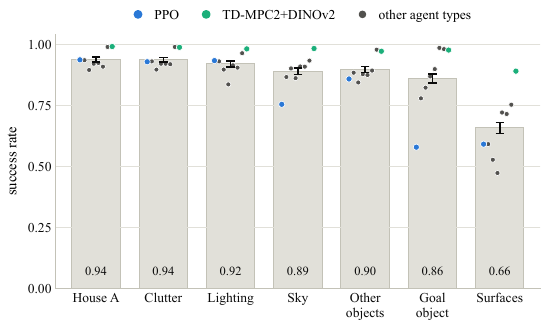}
    \caption{Success rate when one factor changes. Bars: mean over the eight
    agent types (runs that learned house A), with 95\% intervals; dots: each
    agent type. Appendix~\ref{app:single} gives every agent type.}
    \label{fig:single}
  \end{minipage}
\end{figure}

Figure~\ref{fig:reordered} gives the first result. Agents that perform well in
the training house can lose substantial success when the navigation problem is
unchanged and only appearance changes. All agent types are nearly unaffected by
clutter. After lighting, sky, object, and surface changes accumulate, success
ranges from 0.25 for PPO+MAE to 0.83 for TD-MPC2+DINOv2.\footnote{R4 is the
same house as L3, where the separate original-ladder evaluation gives 0.80
(Table~\ref{tab:original}); world-model evaluation is stochastic
(Appendix~\ref{app:ladder}).} PPO declines from 0.94 in A to 0.29 in R4.
TD-MPC2+DINOv2 remains near its training-house rate through R3.

The pooled curve hides meaningful house-to-house variation. In P3, every agent
type reaches at most 0.50 success in the full-shift house, and TD-MPC2+DINOv2
varies from 0.38 to 0.50 across three evaluation passes. In P2, PPO retains
0.79 success in the full-shift house. The benchmark therefore identifies a
general pattern, not a property of every house.

\subsection{Surface materials and goal appearance expose different failures}
\label{sec:surface}

The cumulative ladder establishes that appearance matters. The single-factor
evaluation identifies where that dependence comes from (Table~\ref{tab:single}
and Figure~\ref{fig:single}). Across agent types, changing surfaces lowers mean
success from 0.94 to 0.66. Lighting, sky, non-goal objects, and clutter leave
mean success between 0.89 and 0.94. Surface materials are the lowest-success
single-factor condition for every agent except PPO, where changing only the
goal object produces a comparable result. The surface effect is uneven across
houses: averaged over agent types, the repaint alone lowers success from 0.96
to 0.87 in P0 and from 0.90 to 0.32 in P3 (Table~\ref{tab:single-house}; each
house has one repaint).

That exception separates two kinds of failure. PPO reaches 0.58 when
only the goal appearance changes, against 0.94 in its training house. Changing
only surfaces yields 0.59. The frozen-encoder and world-model agents keep at
least 0.82 success under the goal-object condition, while PPO+Aug, also trained
end to end, falls to 0.78 (Table~\ref{tab:single-agent}). Swapping the goal
asset on top of every other L2 change gives the same pattern
(Table~\ref{tab:goal}): PPO falls from 0.37 to 0.17 and PPO+Aug from 0.38 to
0.23, while the others change by at most 0.09. Geometry does not
explain this: in P0, where the goal surface moves most (0.49\,m), PPO falls
from 0.94 to 0.33 and TD-MPC2 stays at 0.99. End-to-end PPO has thus tied goal
recognition to one asset, while the surface effect spans methods.

The individual effects do not compose linearly. For PPO, the differences from A
caused by the five single changes that make up the full shift (surfaces,
lighting, sky, object appearance, and clutter) sum to $-0.94$, whereas the full
condition changes success by $-0.64$ in the same evaluation. For TD-MPC2, the
individual differences sum to $-0.34$, whereas the full condition changes
success by $-0.45$. Factor-by-factor scores aid diagnosis but do not replace
evaluating the composition.

\subsection{Global image-shift magnitude does not uniquely identify vulnerable factors}
\label{sec:imageshift}

\begin{figure}[t]
  \centering
  \includegraphics[width=0.83\textwidth]{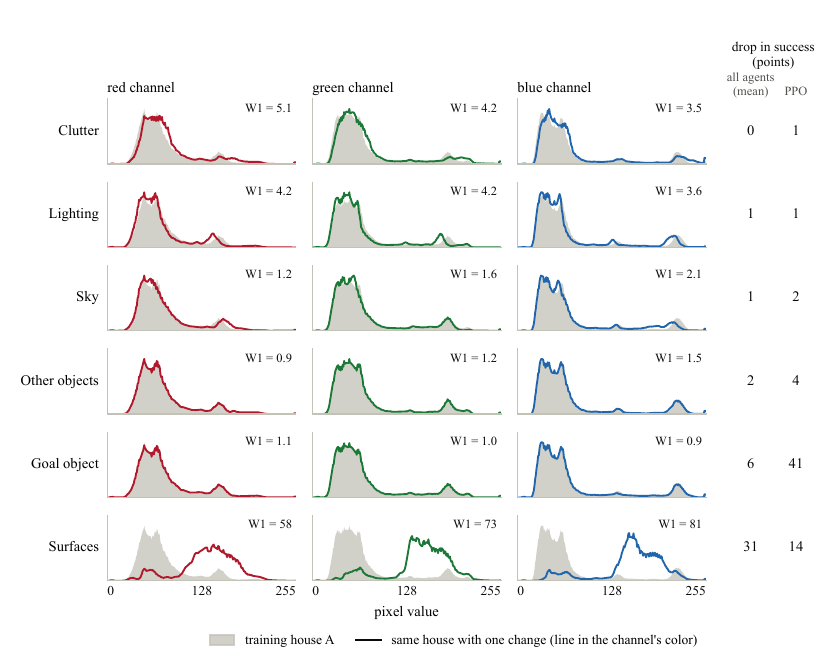}
  \caption{Pixel-value distributions in P1 (Figure~\ref{fig:views}) over the 25
  start views, smoothed over five gray levels. Each row changes one factor.
  Gray: house A; line: the changed house. W1: Wasserstein-1 distance from A in
  gray levels, before smoothing. Right: drop in success in P1, mean of the eight
  agent types and for PPO (runs that learned house A). The goal swap barely
  moves the distributions but costs PPO 41 points; the repaint moves them by
  58--81 gray levels and costs PPO 14.}
  \label{fig:rgb}
\end{figure}

If agents depended mainly on low-level image statistics, larger changes in those
statistics would correspond to lower success. Figure~\ref{fig:rgb} shows the
color distributions behind this measure for each single change in one house.
Across all changes, the drop in success averaged over agent types has a
within-house rank correlation of 0.73 with color-histogram distance
(Appendix~\ref{app:image}). That association is largely induced by surface
repainting, which produces the lowest success and most of the largest color
shifts: 8--71 gray levels, against at most 15 for changes that leave the
surfaces alone. Within each kind, the correlation is 0.37 among changes without
repainting and 0.11 among repaints.

The goal-object control provides a sharper counterexample. In the four houses
with a swappable goal, replacing the goal appearance shifts the color
distribution by about one gray level but changes PPO success from 0.94 to 0.58.
New lighting shifts the distribution by roughly five gray levels and leaves PPO
success at 0.95. Mean per-pixel difference has a higher all-condition
correlation, 0.86, because it retains some information about where the change
occurs.

\begin{table}[t]
  \caption{Success rate along the original cumulative ladder, all 25 runs per
  agent type; mean $\pm$ standard deviation over seeds. Bold: column highest and
  lowest. L1 changes surfaces, lighting, and sky; L2 adds object appearance; L3
  adds clutter. Appendix~\ref{app:ladder} gives every house.}
  \label{tab:original}
  \centering
  \small
  \begin{tabular}{@{}lcccc@{}}
    \toprule
    Agent & A & L1 & L2 & L3 \\
    \midrule
    PPO            & 0.93\,$\pm$\,0.02 & 0.52\,$\pm$\,0.04 & 0.29\,$\pm$\,0.04 & 0.29\,$\pm$\,0.04 \\
    PPO+Aug        & 0.93\,$\pm$\,0.04 & 0.52\,$\pm$\,0.08 & 0.35\,$\pm$\,0.07 & 0.34\,$\pm$\,0.08 \\
    PPO+I-JEPA     & 0.90\,$\pm$\,0.02 & 0.47\,$\pm$\,0.06 & 0.39\,$\pm$\,0.05 & 0.39\,$\pm$\,0.05 \\
    PPO+MAE        & 0.92\,$\pm$\,0.02 & \textbf{0.35}\,$\pm$\,0.08 & \textbf{0.27}\,$\pm$\,0.05 & \textbf{0.26}\,$\pm$\,0.06 \\
    PPO+DINOv2     & 0.92\,$\pm$\,0.04 & 0.72\,$\pm$\,0.09 & 0.65\,$\pm$\,0.06 & 0.66\,$\pm$\,0.07 \\
    DreamerV3      & \textbf{0.76}\,$\pm$\,0.10 & 0.61\,$\pm$\,0.16 & 0.60\,$\pm$\,0.10 & 0.62\,$\pm$\,0.09 \\
    TD-MPC2        & \textbf{0.99}\,$\pm$\,0.00 & 0.64\,$\pm$\,0.09 & 0.54\,$\pm$\,0.09 & 0.54\,$\pm$\,0.09 \\
    TD-MPC2+DINOv2 & \textbf{0.99}\,$\pm$\,0.01 & \textbf{0.88}\,$\pm$\,0.03 & \textbf{0.82}\,$\pm$\,0.06 & \textbf{0.80}\,$\pm$\,0.06 \\
    \bottomrule
  \end{tabular}
\end{table}

Lighting and sky shift the color distribution by similar, small amounts (5.6
and 5.4 gray levels), yet the new sky lowers PPO's success from 0.94 to 0.75
and new lighting leaves it almost unchanged (Appendix~\ref{app:image}).
Neither global measure specifies
whether the changed pixels identify the goal, define surfaces used to navigate,
or belong to irrelevant clutter. An evaluation of planning abstractions should
therefore intervene on task variables rather than rely on an image-distance
proxy.

\subsection{Frozen DINOv2 features improve success under shift, most under TD-MPC2}
\label{sec:pretrained}

On the original ladder (Table~\ref{tab:original}), visual augmentation does not
yield a reliable improvement at the 300k-step budget. PPO and PPO+Aug both reach
0.52 success at L1. At L2, augmentation reaches 0.35 versus 0.29 for PPO, and
the 95\% interval for the difference between the two agents' drops from house A
includes zero (6 points, from $-2$ to 14). At 150k steps, augmentation improves
success at L2, but its advantage decreases after both agents receive more
training (Appendix~\ref{app:budget}).

The frozen encoders differ. PPO+DINOv2 reaches 0.72, 0.65, and 0.66 at L1--L3,
while PPO+I-JEPA and PPO+MAE reach at most 0.47 and 0.35 at L1. Pretraining data
and objective therefore affect which appearance variables the policy depends
on; frozen features are not interchangeable.

TD-MPC2 trained on the same frozen DINOv2 feature scores higher still: 0.88,
0.82, and 0.80 along L1--L3, above both TD-MPC2 with its learned encoder and
PPO+DINOv2. Its single-surface success is 0.89 compared with 0.72 for
PPO+DINOv2. These comparisons do not isolate planning: PPO+DINOv2 and
TD-MPC2+DINOv2 share the single-frame feature but differ in learning algorithm
and trainable size, and TD-MPC2 and TD-MPC2+DINOv2 differ in encoder, image
resolution, and frame history. The agent also remains below its 0.99
training-house success, so the combination reduces rather than removes
dependence on surface appearance.


\section{Discussion}
\label{sec:discussion}

We first state what the evaluation identifies and then what it implies for
future methods.

\subsection{What the controlled evaluation identifies}
\label{sec:identifies}

The experiments distinguish performance under visual shift from the variables
that generate it. Every agent in this study retains some dependence on surface
appearance, and an end-to-end policy also depends on the goal asset. Frozen
DINOv2 features remove much of the sensitivity to lighting, sky, and object
appearance, and TD-MPC2 trained on them reaches higher success still. For that
agent, the remaining failure concentrates on surfaces rather than clutter or
object appearance.

This finding bears on prior results for pretrained features in model-based RL:
\citet{schneider2024pvrmbrl} find that frozen representations need not improve
sample efficiency or out-of-distribution performance, while
\citet{jones2025pvm} find benefits under severe visual shift. Here the same
DINOv2 feature improves both a model-free policy and TD-MPC2. Three
untested differences could account for the contrast with
\citet{schneider2024pvrmbrl}: their factors vary during training, their
simulated scenes differ from our indoor imagery, and their frozen
representation enters through a single linear layer.

\subsection{Requirements for future approaches}
\label{sec:requirements}

\paragraph{Learning where to look.}
Every agent here encodes the whole frame, so a repainted wall filling much of
the view enters the state with the goal. Two results point to selective input:
the goal swap changes few pixels yet lowers PPO's success from 0.94 to 0.58,
and surfaces, which fill most of the view, give the lowest single-factor
success for every agent type except PPO. Foveated primate vision moves a small
high-resolution region to where the task needs
information~\cite{ballard1991animate, hayhoe2005eye}; glimpse policies, patch
attention, and active predictive coding learn such
selection~\cite{mnih2014ram, shang2023active, tang2020neuroevolution,
rao2023apc}. We hypothesize (untested) that learning where to look on strong
pretrained features can close the surface gap.

\paragraph{Counterfactual visual pairs.}
A state encoder can be trained to agree across paired views that preserve
geometry, goal location, and action outcomes while changing surfaces,
illumination, or asset appearance, as our house pairs do. The consistency
objective should be conditioned on the task, to keep the goal identity planning
requires; task-informed objectives are a start~\cite{fu2021tia, zhang2021dbc}.

\paragraph{Objects, geometry, and materials.}
The goal-object result calls for object identity that persists across assets.
The surface result calls for spatial state that represents free space,
boundaries, and navigable layout independently of texture. Object-centric
features, task-conditioned attention, semantic maps, or latent cognitive maps
can provide candidate state variables~\cite{burns2024pvr, george2021cscg}.

\paragraph{Evaluating the next model.}
These are design directions, not methods evaluated here. The benchmark supplies
a direct test: an approach should retain success when the spatial task is
unchanged, identify the goal across assets, and continue to perform when
factors are composed. A world model should also be judged by whether its
planned states and values agree across these interventions.


\section{Limitations}
\label{sec:limitations}

We study one navigation problem in simulation, in five houses, each with a
fixed goal; our conclusions do not extend to ObjectNav with varied goals or
unseen layouts. Agents may partly memorize the goal's location, and surfaces
may cue their own position, so the surface effect may mix lost position cues
with appearance the task does not require. Five houses show substantial
variation that we do not explain, and each has one repaint. We measure success,
not internal states: a drop shows that behavior depends on the changed factor,
not where in the agent, and unchanged success does not show an unchanged state;
comparing latent states across houses at the same poses is the next
measurement. Object swaps change the mesh as well as the texture. The reordered
ladder was chosen from single-factor results on the same agents.
Agent types use each method's own settings and differ in image resolution,
frame input, trainable size, and pretraining (Appendix~\ref{app:settings});
only PPO versus PPO+Aug and PPO+I-JEPA versus PPO+MAE differ in one respect, so
other comparisons between agent types are not ablations. DreamerV3 is likely
undertrained at 300k steps: five of its 25 runs did not learn house A (nine at
150k; Appendix~\ref{app:budget}). Seven of 200 agents were retrained after
saved models were lost, and world-model evaluation is stochastic
(Appendix~\ref{app:ladder}). The reward includes dense straight-line-distance
shaping, and success needs no explicit stop action.


\section{Conclusion}
\label{sec:conclusion}

Our results obtained from a fixed vision-based navigation task reveal that the
behavior of current embodied agents still depends on appearance. Surface
materials account for most of the average reductions in success rate.
Goal-object control reveals an additional asset-specific failure in end-to-end
PPO. Global color-shift magnitude does not determine which intervention harms
an agent. Frozen DINOv2 features, and TD-MPC2 trained on them, increased
navigation success under these changes, but surface appearance remains a
failure mode. Our work offers a new benchmark for future abstraction-learning
methods for embodied planning, evaluating whether they preserve scene geometry
information and goal identity while discarding appearance the task does not
require.


\clearpage
\acknowledgments{This work was supported by AFOSR award no.~FA9550-24-1-0313,
NSF EFRI grant no.~2223495, and a TWCF ``Frameworks'' grant. The work was
facilitated through the use of advanced computational, storage, and networking
infrastructure provided by the UW Hyak supercomputer system.}


\bibliography{references}

@inproceedings{cobbe2020procgen,
  author    = {Karl Cobbe and Christopher Hesse and Jacob Hilton and John Schulman},
  title     = {Leveraging Procedural Generation to Benchmark Reinforcement Learning},
  booktitle = {International Conference on Machine Learning (ICML)},
  year      = {2020}
}

@article{kirk2023survey,
  author  = {Robert Kirk and Amy Zhang and Edward Grefenstette and Tim Rockt{\"a}schel},
  title   = {A Survey of Zero-Shot Generalisation in Deep Reinforcement Learning},
  journal = {Journal of Artificial Intelligence Research},
  volume  = {76},
  pages   = {201--264},
  year    = {2023}
}

@article{stone2021distracting,
  author  = {Austin Stone and Oscar Ramirez and Kurt Konolige and Rico Jonschkowski},
  title   = {The Distracting Control Suite -- A Challenging Benchmark for Reinforcement Learning from Pixels},
  journal = {arXiv preprint arXiv:2101.02722},
  year    = {2021}
}

@inproceedings{hansen2021soda,
  author    = {Nicklas Hansen and Xiaolong Wang},
  title     = {Generalization in Reinforcement Learning by Soft Data Augmentation},
  booktitle = {IEEE International Conference on Robotics and Automation (ICRA)},
  year      = {2021}
}

@inproceedings{yuan2023rlvigen,
  author    = {Zhecheng Yuan and Sizhe Yang and Pu Hua and Can Chang and Kaizhe Hu and Huazhe Xu},
  title     = {{RL-ViGen}: A Reinforcement Learning Benchmark for Visual Generalization},
  booktitle = {Advances in Neural Information Processing Systems (NeurIPS), Datasets and Benchmarks Track},
  year      = {2023}
}

@inproceedings{laskin2020rad,
  author    = {Michael Laskin and Kimin Lee and Adam Stooke and Lerrel Pinto and Pieter Abbeel and Aravind Srinivas},
  title     = {Reinforcement Learning with Augmented Data},
  booktitle = {Advances in Neural Information Processing Systems (NeurIPS)},
  year      = {2020}
}

@inproceedings{yarats2021drq,
  author    = {Denis Yarats and Ilya Kostrikov and Rob Fergus},
  title     = {Image Augmentation Is All You Need: Regularizing Deep Reinforcement Learning from Pixels},
  booktitle = {International Conference on Learning Representations (ICLR)},
  year      = {2021}
}

@inproceedings{yarats2022drqv2,
  author    = {Denis Yarats and Rob Fergus and Alessandro Lazaric and Lerrel Pinto},
  title     = {Mastering Visual Continuous Control: Improved Data-Augmented Reinforcement Learning},
  booktitle = {International Conference on Learning Representations (ICLR)},
  year      = {2022}
}

@inproceedings{deitke2022procthor,
  author    = {Matt Deitke and Eli VanderBilt and Alvaro Herrasti and Luca Weihs and Jordi Salvador and Kiana Ehsani and Winson Han and Eric Kolve and Ali Farhadi and Aniruddha Kembhavi and Roozbeh Mottaghi},
  title     = {{ProcTHOR}: Large-Scale Embodied {AI} Using Procedural Generation},
  booktitle = {Advances in Neural Information Processing Systems (NeurIPS)},
  year      = {2022}
}

@article{kolve2017ai2thor,
  author  = {Eric Kolve and Roozbeh Mottaghi and Winson Han and Eli VanderBilt and Luca Weihs and Alvaro Herrasti and Daniel Gordon and Yuke Zhu and Abhinav Gupta and Ali Farhadi},
  title   = {{AI2-THOR}: An Interactive {3D} Environment for Visual {AI}},
  journal = {arXiv preprint arXiv:1712.05474},
  year    = {2017}
}

@article{schulman2017ppo,
  author  = {John Schulman and Filip Wolski and Prafulla Dhariwal and Alec Radford and Oleg Klimov},
  title   = {Proximal Policy Optimization Algorithms},
  journal = {arXiv preprint arXiv:1707.06347},
  year    = {2017}
}

@article{hafner2025dreamerv3,
  author  = {Danijar Hafner and Jurgis Pasukonis and Jimmy Ba and Timothy Lillicrap},
  title   = {Mastering Diverse Control Tasks through World Models},
  journal = {Nature},
  volume  = {640},
  number  = {8059},
  pages   = {647--653},
  year    = {2025}
}

@inproceedings{hansen2024tdmpc2,
  author    = {Nicklas Hansen and Hao Su and Xiaolong Wang},
  title     = {{TD-MPC2}: Scalable, Robust World Models for Continuous Control},
  booktitle = {International Conference on Learning Representations (ICLR)},
  year      = {2024}
}

@article{oquab2024dinov2,
  author  = {Maxime Oquab and Timoth{\'e}e Darcet and Th{\'e}o Moutakanni and Huy V. Vo and Marc Szafraniec and Vasil Khalidov and Pierre Fernandez and Daniel Haziza and Francisco Massa and Alaaeldin El-Nouby and Mahmoud Assran and Nicolas Ballas and Wojciech Galuba and Russell Howes and Po-Yao Huang and Shang-Wen Li and Ishan Misra and Michael Rabbat and Vasu Sharma and Gabriel Synnaeve and Hu Xu and Herv{\'e} J{\'e}gou and Julien Mairal and Patrick Labatut and Armand Joulin and Piotr Bojanowski},
  title   = {{DINOv2}: Learning Robust Visual Features without Supervision},
  journal = {Transactions on Machine Learning Research},
  year    = {2024}
}

@inproceedings{xie2024decomposing,
  author    = {Annie Xie and Lisa Lee and Ted Xiao and Chelsea Finn},
  title     = {Decomposing the Generalization Gap in Imitation Learning for Visual Robotic Manipulation},
  booktitle = {IEEE International Conference on Robotics and Automation (ICRA)},
  year      = {2024}
}

@inproceedings{gao2024compositional,
  author    = {Jensen Gao and Annie Xie and Ted Xiao and Chelsea Finn and Dorsa Sadigh},
  title     = {Efficient Data Collection for Robotic Manipulation via Compositional Generalization},
  booktitle = {Robotics: Science and Systems (RSS)},
  year      = {2024}
}

@inproceedings{tobin2017domain,
  author    = {Josh Tobin and Rachel Fong and Alex Ray and Jonas Schneider and Wojciech Zaremba and Pieter Abbeel},
  title     = {Domain Randomization for Transferring Deep Neural Networks from Simulation to the Real World},
  booktitle = {IEEE/RSJ International Conference on Intelligent Robots and Systems (IROS)},
  year      = {2017}
}

@inproceedings{parisi2022unsurprising,
  author    = {Simone Parisi and Aravind Rajeswaran and Senthil Purushwalkam and Abhinav Gupta},
  title     = {The Unsurprising Effectiveness of Pre-Trained Vision Models for Control},
  booktitle = {International Conference on Machine Learning (ICML)},
  year      = {2022}
}

@inproceedings{nair2022r3m,
  author    = {Suraj Nair and Aravind Rajeswaran and Vikash Kumar and Chelsea Finn and Abhinav Gupta},
  title     = {{R3M}: A Universal Visual Representation for Robot Manipulation},
  booktitle = {Conference on Robot Learning (CoRL)},
  year      = {2022}
}

@inproceedings{majumdar2023vc1,
  author    = {Arjun Majumdar and Karmesh Yadav and Sergio Arnaud and Yecheng Jason Ma and Claire Chen and Sneha Silwal and Aryan Jain and Vincent-Pierre Berges and Tingfan Wu and Jay Vakil and Pieter Abbeel and Jitendra Malik and Dhruv Batra and Yixin Lin and Oleksandr Maksymets and Aravind Rajeswaran and Franziska Meier},
  title     = {Where Are We in the Search for an Artificial Visual Cortex for Embodied Intelligence?},
  booktitle = {Advances in Neural Information Processing Systems (NeurIPS)},
  year      = {2023}
}

@inproceedings{hansen2023pretraining,
  author    = {Nicklas Hansen and Zhecheng Yuan and Yanjie Ze and Tongzhou Mu and Aravind Rajeswaran and Hao Su and Huazhe Xu and Xiaolong Wang},
  title     = {On Pre-Training for Visuo-Motor Control: Revisiting a Learning-from-Scratch Baseline},
  booktitle = {International Conference on Machine Learning (ICML)},
  year      = {2023}
}

@inproceedings{burns2024pvr,
  author    = {Kaylee Burns and Zach Witzel and Jubayer Ibn Hamid and Tianhe Yu and Chelsea Finn and Karol Hausman},
  title     = {What Makes Pre-Trained Visual Representations Successful for Robust Manipulation?},
  booktitle = {Conference on Robot Learning (CoRL)},
  year      = {2024}
}

@inproceedings{schneider2024pvrmbrl,
  author    = {Moritz Schneider and Robert Krug and Narunas Vaskevicius and Luigi Palmieri and Joschka Boedecker},
  title     = {The Surprising Ineffectiveness of Pre-Trained Visual Representations for Model-Based Reinforcement Learning},
  booktitle = {Advances in Neural Information Processing Systems (NeurIPS)},
  year      = {2024}
}

@article{jones2025pvm,
  author  = {S. Jones and L. Zhou and S. W. Pattinson},
  title   = {Pre-Trained Visual Representations Generalize Where It Matters in Model-Based Reinforcement Learning},
  journal = {arXiv preprint arXiv:2509.12531},
  year    = {2025}
}

@inproceedings{fu2021tia,
  author    = {Xiang Fu and Ge Yang and Pulkit Agrawal and Tommi Jaakkola},
  title     = {Learning Task Informed Abstractions},
  booktitle = {International Conference on Machine Learning (ICML)},
  year      = {2021}
}

@inproceedings{zhang2021dbc,
  author    = {Amy Zhang and Rowan McAllister and Roberto Calandra and Yarin Gal and Sergey Levine},
  title     = {Learning Invariant Representations for Reinforcement Learning without Reconstruction},
  booktitle = {International Conference on Learning Representations (ICLR)},
  year      = {2021}
}

@article{batra2020objectnav,
  author  = {Dhruv Batra and Aaron Gokaslan and Aniruddha Kembhavi and Oleksandr Maksymets and Roozbeh Mottaghi and Manolis Savva and Alexander Toshev and Erik Wijmans},
  title   = {{ObjectNav} Revisited: On Evaluation of Embodied Agents Navigating to Objects},
  journal = {arXiv preprint arXiv:2006.13171},
  year    = {2020}
}

@article{anderson2018evaluation,
  author  = {Peter Anderson and Angel Chang and Devendra Singh Chaplot and Alexey Dosovitskiy and Saurabh Gupta and Vladlen Koltun and Jana Kosecka and Jitendra Malik and Roozbeh Mottaghi and Manolis Savva and Amir R. Zamir},
  title   = {On Evaluation of Embodied Navigation Agents},
  journal = {arXiv preprint arXiv:1807.06757},
  year    = {2018}
}

@inproceedings{agarwal2021precipice,
  author    = {Rishabh Agarwal and Max Schwarzer and Pablo Samuel Castro and Aaron Courville and Marc G. Bellemare},
  title     = {Deep Reinforcement Learning at the Edge of the Statistical Precipice},
  booktitle = {Advances in Neural Information Processing Systems (NeurIPS)},
  year      = {2021}
}

@article{mnih2015dqn,
  author  = {Volodymyr Mnih and Koray Kavukcuoglu and David Silver and Andrei A. Rusu and Joel Veness and Marc G. Bellemare and Alex Graves and Martin Riedmiller and Andreas K. Fidjeland and Georg Ostrovski and Stig Petersen and Charles Beattie and Amir Sadik and Ioannis Antonoglou and Helen King and Dharshan Kumaran and Daan Wierstra and Shane Legg and Demis Hassabis},
  title   = {Human-Level Control through Deep Reinforcement Learning},
  journal = {Nature},
  volume  = {518},
  number  = {7540},
  pages   = {529--533},
  year    = {2015}
}

@article{raffin2021sb3,
  author  = {Antonin Raffin and Ashley Hill and Adam Gleave and Anssi Kanervisto and Maximilian Ernestus and Noah Dormann},
  title   = {{Stable-Baselines3}: Reliable Reinforcement Learning Implementations},
  journal = {Journal of Machine Learning Research},
  volume  = {22},
  number  = {268},
  pages   = {1--8},
  year    = {2021}
}

@inproceedings{assran2023ijepa,
  author    = {Mahmoud Assran and Quentin Duval and Ishan Misra and Piotr Bojanowski and Pascal Vincent and Michael Rabbat and Yann LeCun and Nicolas Ballas},
  title     = {Self-Supervised Learning from Images with a Joint-Embedding Predictive Architecture},
  booktitle = {IEEE/CVF Conference on Computer Vision and Pattern Recognition (CVPR)},
  year      = {2023}
}

@inproceedings{he2022mae,
  author    = {Kaiming He and Xinlei Chen and Saining Xie and Yanghao Li and Piotr Doll{\'a}r and Ross Girshick},
  title     = {Masked Autoencoders Are Scalable Vision Learners},
  booktitle = {IEEE/CVF Conference on Computer Vision and Pattern Recognition (CVPR)},
  year      = {2022}
}

@article{ballard1991animate,
  author  = {Dana H. Ballard},
  title   = {Animate Vision},
  journal = {Artificial Intelligence},
  volume  = {48},
  number  = {1},
  pages   = {57--86},
  year    = {1991}
}

@article{hayhoe2005eye,
  author  = {Mary Hayhoe and Dana Ballard},
  title   = {Eye Movements in Natural Behavior},
  journal = {Trends in Cognitive Sciences},
  volume  = {9},
  number  = {4},
  pages   = {188--194},
  year    = {2005}
}

@inproceedings{mnih2014ram,
  author    = {Volodymyr Mnih and Nicolas Heess and Alex Graves and Koray Kavukcuoglu},
  title     = {Recurrent Models of Visual Attention},
  booktitle = {Advances in Neural Information Processing Systems (NeurIPS)},
  year      = {2014}
}

@inproceedings{shang2023active,
  author    = {Jinghuan Shang and Michael S. Ryoo},
  title     = {Active Vision Reinforcement Learning under Limited Visual Observability},
  booktitle = {Advances in Neural Information Processing Systems (NeurIPS)},
  year      = {2023}
}

@inproceedings{tang2020neuroevolution,
  author    = {Yujin Tang and Duong Nguyen and David Ha},
  title     = {Neuroevolution of Self-Interpretable Agents},
  booktitle = {Genetic and Evolutionary Computation Conference (GECCO)},
  year      = {2020}
}

@article{rao2023apc,
  author  = {Rajesh P. N. Rao and Dimitrios C. Gklezakos and Vishwas Sathish},
  title   = {Active Predictive Coding: A Unifying Neural Model for Active Perception, Compositional Learning, and Hierarchical Planning},
  journal = {Neural Computation},
  volume  = {36},
  number  = {1},
  pages   = {1--32},
  year    = {2023}
}

@article{george2021cscg,
  author  = {Dileep George and Rajeev V. Rikhye and Nishad Gothoskar and J. Swaroop Guntupalli and Antoine Dedieu and Miguel L{\'a}zaro-Gredilla},
  title   = {Clone-Structured Graph Representations Enable Flexible Learning and Vicarious Evaluation of Cognitive Maps},
  journal = {Nature Communications},
  volume  = {12},
  number  = {1},
  pages   = {2392},
  year    = {2021}
}

@article{holm1979,
  author  = {Sture Holm},
  title   = {A Simple Sequentially Rejective Multiple Test Procedure},
  journal = {Scandinavian Journal of Statistics},
  volume  = {6},
  number  = {2},
  pages   = {65--70},
  year    = {1979}
}

@inproceedings{pumacay2024colosseum,
  author    = {Wilbert Pumacay and Ishika Singh and Jiafei Duan and Ranjay Krishna and Jesse Thomason and Dieter Fox},
  title     = {{THE COLOSSEUM}: A Benchmark for Evaluating Generalization for Robotic Manipulation},
  booktitle = {Robotics: Science and Systems (RSS)},
  year      = {2024}
}

@inproceedings{xing2021kitchenshift,
  author    = {Eliot Xing and Abhinav Gupta and Samantha Powers and Victoria Dean},
  title     = {{KitchenShift}: Evaluating Zero-Shot Generalization of Imitation-Based Policy Learning Under Domain Shifts},
  booktitle = {NeurIPS 2021 Workshop on Distribution Shifts: Connecting Methods and Applications},
  year      = {2021}
}


\clearpage
\appendix

\section{House pairs and verification}
\label{app:houses}

Every house in the two ladders and the single-factor evaluation passed the three
checks of Section~\ref{sec:interventions} on three independent simulator loads.
The goal-control houses (L2noT, Appendix~\ref{app:goal}) passed them on one
load. Table~\ref{tab:houses} lists, for each house pair, the goal object, the
number of reachable positions, the number of clutter objects, whether the goal
asset could be swapped, and the largest change in path length to the goal. In
the single-factor houses, surface, lighting, sky, and clutter changes leave
that path length unchanged (0.00\,m in all five pairs); swapping the non-goal
objects changes it by at most 0.07\,m, and swapping the goal by 0.25, 0.49,
0.25, and 0.00\,m in P1, P0, P3, and P4.

\begin{table}[ht]
  \caption{House-pair checks. The path-shift columns give the largest change in
  geodesic distance from an evaluation start pose to the goal surface relative
  to house A. SPL uses house-A distances throughout.}
  \label{tab:houses}
  \centering
  \small
  \resizebox{\textwidth}{!}{%
  \begin{tabular}{@{}lcccccc@{}}
    \toprule
    Pair & Target & Reachable cells & Clutter kept & Goal swapped & L1 path shift (m) & L2/L3 path shift (m) \\
    \midrule
    P2 & Bed        & 128 & 5 & no  & 0.00 & 0.00 / 0.00 \\
    P1 & Bed        & 162 & 2 & yes & 0.00 & 0.25 / 0.25 \\
    P0 & Fridge     & 380 & 4 & yes & 0.00 & 0.49 / 0.49 \\
    P3 & Bed        & 460 & 8 & yes & 0.00 & 0.26 / 0.26 \\
    P4 & Television & 681 & 8 & yes & 0.00 & 0.00 / 0.00 \\
    \bottomrule
  \end{tabular}
  }
\end{table}

\section{Agent settings}
\label{app:settings}

Table~\ref{tab:settings} lists the training settings of every agent type. Each
agent type uses its method's own settings, so the types differ in several
respects besides the one each is named for. Image resolution: the PPO agents
receive $128{\times}128$ images, resized to $224{\times}224$ for the frozen
encoders, while DreamerV3 and TD-MPC2 receive $64{\times}64$ images and
TD-MPC2+DINOv2 receives the PPO+DINOv2 feature. Frame
input: TD-MPC2 stacks three frames, DreamerV3 carries a recurrent state, and
the other agents receive one frame. Trainable size: the frozen-encoder PPO
agents train a head of 0.17M (I-JEPA, MAE) or 0.21M (DINOv2) parameters, while
end-to-end PPO trains a 4.8M-parameter encoder and head. Pretraining: I-JEPA
and MAE use the same ViT-H/14 architecture and differ in pretraining objective,
while DINOv2 uses a larger ViT-g/14 pretrained on a larger image collection.
PPO versus PPO+Aug (augmentation) and PPO+I-JEPA versus PPO+MAE (pretraining
objective) are the only pairs that differ in a single respect.

\begin{table}[ht]
  \caption{Training settings. All agents use 300k environment steps, 200-step
  episodes, and the same 25 pinned evaluation start poses.}
  \label{tab:settings}
  \centering
  \small
  \begin{tabular}{@{}l>{\raggedright\arraybackslash}p{0.72\textwidth}@{}}
    \toprule
    Agent & Settings \\
    \midrule
    PPO & Stable-Baselines3; Nature-CNN encoder~\cite{mnih2015dqn} on
    $128{\times}128$ RGB; rollout 1024 steps, minibatch 128, 4 epochs, learning
    rate $2.5{\times}10^{-4}$, $\gamma{=}0.99$, generalized advantage estimation
    (GAE) $\lambda{=}0.95$, clip 0.2, entropy coefficient 0.01. \\
    PPO+Aug & As PPO; brightness, contrast, and saturation jitter $\pm0.4$, hue
    $\pm36^\circ$, one draw per episode during training only. \\
    PPO+frozen & As PPO, with the encoder replaced by frozen I-JEPA ViT-H/14,
    MAE ViT-H/14, or DINOv2 ViT-g/14; bilinear resize to $224{\times}224$,
    ImageNet normalization, mean patch tokens with class and register tokens
    removed, half-precision (fp16) inference, and a two-layer 64-unit MLP head.
    MAE runs without masking. \\
    DreamerV3 & PyTorch reimplementation
    (\url{https://github.com/NM512/dreamerv3-torch}), default small model;
    $64{\times}64$ observations; replay ratio 512; batch $16{\times}64$; 2,500
    random prefill steps. \\
    TD-MPC2 & Official implementation
    (\url{https://github.com/nicklashansen/tdmpc2}), 5M-parameter model;
    $64{\times}64$ observations, three-frame stack; horizon 3; MPPI with 6
    iterations, 512 samples, 64 elites; batch 256; one update per environment
    step; discrete action by argmax over a five-dimensional continuous output. \\
    TD-MPC2+DINOv2 & As TD-MPC2, with the frozen 1536-dimensional DINOv2
    ViT-g/14 feature used by PPO+DINOv2 as its one-frame observation, entering
    through TD-MPC2's MLP state encoder. \\
    \bottomrule
  \end{tabular}
\end{table}

\section{Cumulative-ladder success rates}
\label{app:ladder}

\begin{figure}[ht]
  \centering
  \includegraphics{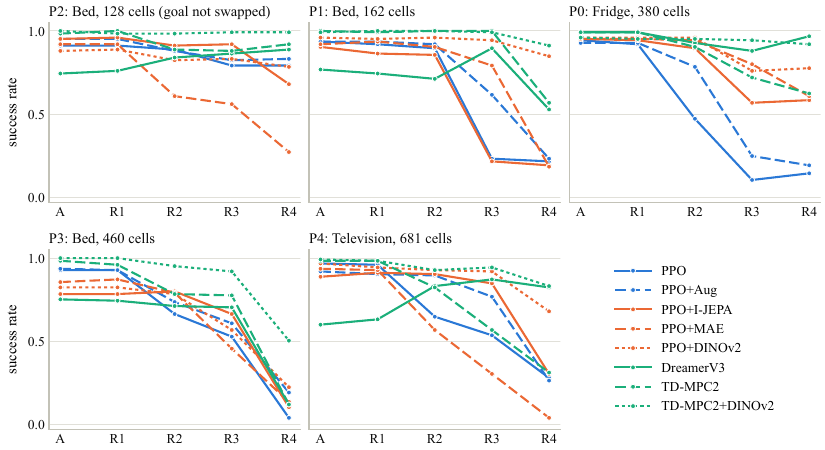}
  \caption{Success rate along the reordered cumulative ladder in each house,
  mean over the five training seeds, every run.}
  \label{fig:by-house}
\end{figure}

Table~\ref{tab:reordered} gives the reordered ladder and
Table~\ref{tab:original} the original ladder; Figure~\ref{fig:by-house} and
Tables~\ref{tab:reordered-house} and~\ref{tab:original-house} give every house.
All values are mean $\pm$ standard deviation over the five training seeds, and
bold marks the highest and lowest value in each column (in the per-house
tables, within each house). The full-shift house (L3, which is also R4) was
evaluated in three separate passes (Table~\ref{tab:passes}). The PPO-based
agents agree across passes to within 0.02 on all five houses pooled. The world
models vary more: their evaluation is stochastic, and the two later passes use
retrained models for seven cells whose original models were lost.

\begin{table}[ht]
  \caption{Success rate along the reordered cumulative ladder, every agent,
  every run; mean $\pm$ standard deviation over the five training seeds. Bold
  marks the highest and lowest value in each column.}
  \label{tab:reordered}
  \centering
  \scriptsize
  \resizebox{\textwidth}{!}{%
  \begin{tabular}{@{}lccccc@{}}
    \toprule
    Agent & House A & R1 & R2 & R3 & R4 \\
          &         & clutter & + lighting \& sky & + object looks & + surfaces \\
    \midrule
    \multicolumn{6}{@{}l}{\textit{Model-free, encoder trained from scratch}} \\
    \quad PPO     & 0.94\,$\pm$\,0.03 & 0.93\,$\pm$\,0.03 & \textbf{0.71}\,$\pm$\,0.07 & \textbf{0.44}\,$\pm$\,0.04 & 0.29\,$\pm$\,0.04 \\
    \quad PPO+Aug & 0.93\,$\pm$\,0.04 & 0.93\,$\pm$\,0.03 & 0.84\,$\pm$\,0.06 & 0.61\,$\pm$\,0.08 & 0.34\,$\pm$\,0.08 \\
    \addlinespace[2pt]
    \multicolumn{6}{@{}l}{\textit{Model-free, frozen pretrained encoder}} \\
    \quad PPO+I-JEPA & 0.90\,$\pm$\,0.02 & 0.89\,$\pm$\,0.02 & 0.87\,$\pm$\,0.02 & 0.64\,$\pm$\,0.02 & 0.37\,$\pm$\,0.05 \\
    \quad PPO+MAE    & 0.92\,$\pm$\,0.02 & 0.92\,$\pm$\,0.02 & 0.76\,$\pm$\,0.04 & 0.58\,$\pm$\,0.11 & \textbf{0.25}\,$\pm$\,0.06 \\
    \quad PPO+DINOv2 & 0.92\,$\pm$\,0.04 & 0.91\,$\pm$\,0.04 & 0.89\,$\pm$\,0.05 & 0.80\,$\pm$\,0.04 & 0.66\,$\pm$\,0.07 \\
    \addlinespace[2pt]
    \multicolumn{6}{@{}l}{\textit{World model, encoder trained from scratch}} \\
    \quad DreamerV3 & \textbf{0.77}\,$\pm$\,0.09 & \textbf{0.77}\,$\pm$\,0.08 & 0.80\,$\pm$\,0.11 & 0.84\,$\pm$\,0.14 & 0.67\,$\pm$\,0.08 \\
    \quad TD-MPC2   & \textbf{0.99}\,$\pm$\,0.01 & \textbf{0.99}\,$\pm$\,0.01 & 0.88\,$\pm$\,0.03 & 0.79\,$\pm$\,0.05 & 0.51\,$\pm$\,0.08 \\
    \addlinespace[2pt]
    \multicolumn{6}{@{}l}{\textit{World model, frozen pretrained encoder}} \\
    \quad TD-MPC2+DINOv2 & \textbf{0.99}\,$\pm$\,0.00 & 0.98\,$\pm$\,0.01 & \textbf{0.96}\,$\pm$\,0.02 & \textbf{0.96}\,$\pm$\,0.02 & \textbf{0.83}\,$\pm$\,0.06 \\
    \bottomrule
  \end{tabular}
  }
\end{table}

\begin{table}[ht]
  \caption{Success rate in each house along the reordered cumulative ladder,
  every run; mean $\pm$ standard deviation over the five training seeds. Bold
  marks the highest and lowest value in each column within each house.}
  \label{tab:reordered-house}
  \centering
  \scriptsize
  \begin{tabular}{@{}llccccc@{}}
    \toprule
    Pair & Agent & A & R1 & R2 & R3 & R4 \\
    \midrule
    \multirow{8}{*}{\shortstack[l]{P2\\Bed}}
      & PPO            & 0.91\,$\pm$\,0.03 & 0.91\,$\pm$\,0.03 & 0.89\,$\pm$\,0.03 & 0.79\,$\pm$\,0.07 & 0.79\,$\pm$\,0.03 \\
      & PPO+Aug        & 0.95\,$\pm$\,0.05 & 0.95\,$\pm$\,0.05 & 0.88\,$\pm$\,0.03 & 0.82\,$\pm$\,0.07 & 0.83\,$\pm$\,0.07 \\
      & PPO+I-JEPA     & 0.95\,$\pm$\,0.07 & 0.96\,$\pm$\,0.07 & 0.91\,$\pm$\,0.07 & 0.92\,$\pm$\,0.08 & 0.68\,$\pm$\,0.20 \\
      & PPO+MAE        & 0.92\,$\pm$\,0.12 & 0.92\,$\pm$\,0.12 & \textbf{0.61}\,$\pm$\,0.29 & \textbf{0.56}\,$\pm$\,0.26 & \textbf{0.27}\,$\pm$\,0.30 \\
      & PPO+DINOv2     & 0.88\,$\pm$\,0.06 & 0.89\,$\pm$\,0.06 & 0.82\,$\pm$\,0.10 & 0.83\,$\pm$\,0.13 & 0.78\,$\pm$\,0.09 \\
      & DreamerV3      & \textbf{0.74}\,$\pm$\,0.39 & \textbf{0.76}\,$\pm$\,0.38 & 0.84\,$\pm$\,0.27 & 0.86\,$\pm$\,0.20 & 0.89\,$\pm$\,0.12 \\
      & TD-MPC2        & 0.98\,$\pm$\,0.02 & \textbf{1.00}\,$\pm$\,0.00 & 0.89\,$\pm$\,0.07 & 0.88\,$\pm$\,0.03 & 0.92\,$\pm$\,0.06 \\
      & TD-MPC2+DINOv2 & \textbf{1.00}\,$\pm$\,0.00 & 0.98\,$\pm$\,0.02 & \textbf{0.98}\,$\pm$\,0.02 & \textbf{0.99}\,$\pm$\,0.02 & \textbf{0.99}\,$\pm$\,0.02 \\
    \midrule
    \multirow{8}{*}{\shortstack[l]{P1\\Bed}}
      & PPO            & 0.94\,$\pm$\,0.05 & 0.92\,$\pm$\,0.06 & 0.90\,$\pm$\,0.04 & 0.23\,$\pm$\,0.07 & 0.22\,$\pm$\,0.09 \\
      & PPO+Aug        & 0.94\,$\pm$\,0.06 & 0.94\,$\pm$\,0.05 & 0.92\,$\pm$\,0.06 & 0.62\,$\pm$\,0.27 & 0.23\,$\pm$\,0.10 \\
      & PPO+I-JEPA     & 0.90\,$\pm$\,0.02 & 0.86\,$\pm$\,0.04 & 0.86\,$\pm$\,0.05 & \textbf{0.22}\,$\pm$\,0.06 & 0.19\,$\pm$\,0.02 \\
      & PPO+MAE        & 0.92\,$\pm$\,0.03 & 0.94\,$\pm$\,0.02 & 0.90\,$\pm$\,0.02 & 0.79\,$\pm$\,0.05 & \textbf{0.18}\,$\pm$\,0.02 \\
      & PPO+DINOv2     & 0.96\,$\pm$\,0.05 & 0.95\,$\pm$\,0.05 & 0.96\,$\pm$\,0.07 & 0.94\,$\pm$\,0.05 & 0.85\,$\pm$\,0.06 \\
      & DreamerV3      & \textbf{0.77}\,$\pm$\,0.27 & \textbf{0.74}\,$\pm$\,0.29 & \textbf{0.71}\,$\pm$\,0.33 & 0.90\,$\pm$\,0.23 & 0.53\,$\pm$\,0.27 \\
      & TD-MPC2        & \textbf{1.00}\,$\pm$\,0.00 & 0.99\,$\pm$\,0.02 & \textbf{1.00}\,$\pm$\,0.00 & \textbf{1.00}\,$\pm$\,0.00 & 0.57\,$\pm$\,0.26 \\
      & TD-MPC2+DINOv2 & 0.99\,$\pm$\,0.02 & \textbf{1.00}\,$\pm$\,0.00 & \textbf{1.00}\,$\pm$\,0.00 & 0.99\,$\pm$\,0.02 & \textbf{0.91}\,$\pm$\,0.13 \\
    \midrule
    \multirow{8}{*}{\shortstack[l]{P0\\Fridge}}
      & PPO            & 0.94\,$\pm$\,0.07 & \textbf{0.92}\,$\pm$\,0.07 & \textbf{0.47}\,$\pm$\,0.23 & \textbf{0.10}\,$\pm$\,0.06 & \textbf{0.14}\,$\pm$\,0.10 \\
      & PPO+Aug        & \textbf{0.93}\,$\pm$\,0.05 & 0.93\,$\pm$\,0.05 & 0.78\,$\pm$\,0.15 & 0.25\,$\pm$\,0.17 & 0.19\,$\pm$\,0.06 \\
      & PPO+I-JEPA     & 0.95\,$\pm$\,0.02 & 0.94\,$\pm$\,0.04 & 0.90\,$\pm$\,0.02 & 0.57\,$\pm$\,0.15 & 0.58\,$\pm$\,0.15 \\
      & PPO+MAE        & 0.95\,$\pm$\,0.04 & 0.95\,$\pm$\,0.04 & 0.94\,$\pm$\,0.02 & 0.80\,$\pm$\,0.18 & 0.61\,$\pm$\,0.11 \\
      & PPO+DINOv2     & 0.96\,$\pm$\,0.03 & 0.96\,$\pm$\,0.03 & \textbf{0.96}\,$\pm$\,0.04 & 0.76\,$\pm$\,0.06 & 0.78\,$\pm$\,0.09 \\
      & DreamerV3      & \textbf{0.99}\,$\pm$\,0.02 & \textbf{0.99}\,$\pm$\,0.02 & 0.93\,$\pm$\,0.10 & 0.88\,$\pm$\,0.11 & \textbf{0.97}\,$\pm$\,0.04 \\
      & TD-MPC2        & \textbf{0.99}\,$\pm$\,0.02 & \textbf{0.99}\,$\pm$\,0.02 & 0.90\,$\pm$\,0.08 & 0.72\,$\pm$\,0.09 & 0.62\,$\pm$\,0.16 \\
      & TD-MPC2+DINOv2 & 0.96\,$\pm$\,0.00 & 0.95\,$\pm$\,0.02 & 0.95\,$\pm$\,0.02 & \textbf{0.94}\,$\pm$\,0.04 & 0.92\,$\pm$\,0.03 \\
    \midrule
    \multirow{8}{*}{\shortstack[l]{P3\\Bed}}
      & PPO            & 0.93\,$\pm$\,0.06 & 0.93\,$\pm$\,0.06 & \textbf{0.66}\,$\pm$\,0.21 & 0.53\,$\pm$\,0.18 & \textbf{0.04}\,$\pm$\,0.05 \\
      & PPO+Aug        & 0.94\,$\pm$\,0.08 & 0.93\,$\pm$\,0.08 & 0.74\,$\pm$\,0.13 & 0.61\,$\pm$\,0.11 & 0.19\,$\pm$\,0.18 \\
      & PPO+I-JEPA     & 0.78\,$\pm$\,0.07 & 0.78\,$\pm$\,0.06 & 0.80\,$\pm$\,0.06 & 0.66\,$\pm$\,0.05 & 0.10\,$\pm$\,0.14 \\
      & PPO+MAE        & 0.86\,$\pm$\,0.07 & 0.87\,$\pm$\,0.09 & 0.79\,$\pm$\,0.11 & \textbf{0.46}\,$\pm$\,0.24 & 0.14\,$\pm$\,0.20 \\
      & PPO+DINOv2     & 0.82\,$\pm$\,0.18 & 0.82\,$\pm$\,0.18 & 0.78\,$\pm$\,0.20 & 0.57\,$\pm$\,0.15 & 0.22\,$\pm$\,0.24 \\
      & DreamerV3      & \textbf{0.75}\,$\pm$\,0.43 & \textbf{0.74}\,$\pm$\,0.35 & 0.71\,$\pm$\,0.35 & 0.70\,$\pm$\,0.32 & 0.12\,$\pm$\,0.05 \\
      & TD-MPC2        & 0.98\,$\pm$\,0.02 & 0.96\,$\pm$\,0.03 & 0.78\,$\pm$\,0.08 & 0.78\,$\pm$\,0.18 & 0.12\,$\pm$\,0.10 \\
      & TD-MPC2+DINOv2 & \textbf{1.00}\,$\pm$\,0.00 & \textbf{1.00}\,$\pm$\,0.00 & \textbf{0.95}\,$\pm$\,0.07 & \textbf{0.92}\,$\pm$\,0.07 & \textbf{0.50}\,$\pm$\,0.16 \\
    \midrule
    \multirow{8}{*}{\shortstack[l]{P4\\Television}}
      & PPO            & 0.97\,$\pm$\,0.04 & 0.96\,$\pm$\,0.04 & 0.65\,$\pm$\,0.28 & 0.54\,$\pm$\,0.24 & 0.28\,$\pm$\,0.23 \\
      & PPO+Aug        & 0.92\,$\pm$\,0.07 & 0.90\,$\pm$\,0.11 & 0.90\,$\pm$\,0.09 & 0.77\,$\pm$\,0.17 & 0.26\,$\pm$\,0.25 \\
      & PPO+I-JEPA     & 0.89\,$\pm$\,0.05 & 0.91\,$\pm$\,0.03 & 0.90\,$\pm$\,0.05 & 0.85\,$\pm$\,0.05 & 0.30\,$\pm$\,0.25 \\
      & PPO+MAE        & 0.94\,$\pm$\,0.02 & 0.93\,$\pm$\,0.03 & \textbf{0.57}\,$\pm$\,0.25 & \textbf{0.30}\,$\pm$\,0.31 & \textbf{0.04}\,$\pm$\,0.09 \\
      & PPO+DINOv2     & 0.97\,$\pm$\,0.03 & 0.94\,$\pm$\,0.02 & \textbf{0.93}\,$\pm$\,0.07 & 0.92\,$\pm$\,0.08 & 0.68\,$\pm$\,0.24 \\
      & DreamerV3      & \textbf{0.60}\,$\pm$\,0.43 & \textbf{0.63}\,$\pm$\,0.40 & 0.83\,$\pm$\,0.17 & 0.87\,$\pm$\,0.13 & 0.82\,$\pm$\,0.19 \\
      & TD-MPC2        & 0.98\,$\pm$\,0.02 & \textbf{0.98}\,$\pm$\,0.02 & 0.82\,$\pm$\,0.07 & 0.57\,$\pm$\,0.23 & 0.31\,$\pm$\,0.23 \\
      & TD-MPC2+DINOv2 & \textbf{0.99}\,$\pm$\,0.02 & \textbf{0.98}\,$\pm$\,0.04 & \textbf{0.93}\,$\pm$\,0.06 & \textbf{0.94}\,$\pm$\,0.02 & \textbf{0.83}\,$\pm$\,0.10 \\
    \bottomrule
  \end{tabular}
\end{table}

\begin{table}[ht]
  \caption{Success rate in each house along the original cumulative ladder,
  every run; mean $\pm$ standard deviation over the five training seeds. Bold
  marks the highest and lowest value in each column within each house.}
  \label{tab:original-house}
  \centering
  \scriptsize
  \begin{tabular}{@{}llcccc@{}}
    \toprule
    Pair & Agent & A & L1 & L2 & L3 \\
    \midrule
    \multirow{8}{*}{\shortstack[l]{P2\\Bed}}
      & PPO            & 0.91\,$\pm$\,0.03 & 0.86\,$\pm$\,0.07 & 0.79\,$\pm$\,0.03 & 0.79\,$\pm$\,0.03 \\
      & PPO+Aug        & 0.95\,$\pm$\,0.05 & 0.86\,$\pm$\,0.05 & 0.83\,$\pm$\,0.07 & 0.83\,$\pm$\,0.07 \\
      & PPO+I-JEPA     & 0.96\,$\pm$\,0.07 & 0.62\,$\pm$\,0.28 & 0.70\,$\pm$\,0.18 & 0.70\,$\pm$\,0.18 \\
      & PPO+MAE        & 0.92\,$\pm$\,0.12 & \textbf{0.32}\,$\pm$\,0.38 & \textbf{0.26}\,$\pm$\,0.29 & \textbf{0.27}\,$\pm$\,0.30 \\
      & PPO+DINOv2     & 0.90\,$\pm$\,0.02 & 0.77\,$\pm$\,0.09 & 0.76\,$\pm$\,0.10 & 0.78\,$\pm$\,0.09 \\
      & DreamerV3      & \textbf{0.76}\,$\pm$\,0.39 & 0.72\,$\pm$\,0.37 & 0.69\,$\pm$\,0.39 & 0.70\,$\pm$\,0.39 \\
      & TD-MPC2        & \textbf{1.00}\,$\pm$\,0.00 & 0.93\,$\pm$\,0.09 & 0.92\,$\pm$\,0.07 & 0.94\,$\pm$\,0.08 \\
      & TD-MPC2+DINOv2 & \textbf{1.00}\,$\pm$\,0.00 & \textbf{0.98}\,$\pm$\,0.04 & \textbf{0.99}\,$\pm$\,0.02 & \textbf{1.00}\,$\pm$\,0.00 \\
    \midrule
    \multirow{8}{*}{\shortstack[l]{P1\\Bed}}
      & PPO            & 0.93\,$\pm$\,0.05 & 0.62\,$\pm$\,0.16 & 0.22\,$\pm$\,0.09 & 0.22\,$\pm$\,0.09 \\
      & PPO+Aug        & 0.94\,$\pm$\,0.06 & 0.46\,$\pm$\,0.28 & 0.30\,$\pm$\,0.13 & 0.23\,$\pm$\,0.10 \\
      & PPO+I-JEPA     & 0.90\,$\pm$\,0.02 & \textbf{0.23}\,$\pm$\,0.05 & 0.19\,$\pm$\,0.02 & 0.19\,$\pm$\,0.02 \\
      & PPO+MAE        & 0.92\,$\pm$\,0.03 & 0.26\,$\pm$\,0.06 & \textbf{0.17}\,$\pm$\,0.05 & \textbf{0.18}\,$\pm$\,0.02 \\
      & PPO+DINOv2     & 0.94\,$\pm$\,0.05 & 0.86\,$\pm$\,0.08 & 0.81\,$\pm$\,0.09 & 0.82\,$\pm$\,0.05 \\
      & DreamerV3      & \textbf{0.50}\,$\pm$\,0.22 & 0.56\,$\pm$\,0.30 & 0.43\,$\pm$\,0.16 & 0.55\,$\pm$\,0.18 \\
      & TD-MPC2        & \textbf{1.00}\,$\pm$\,0.00 & 0.60\,$\pm$\,0.24 & 0.62\,$\pm$\,0.29 & 0.57\,$\pm$\,0.19 \\
      & TD-MPC2+DINOv2 & 0.99\,$\pm$\,0.02 & \textbf{0.96}\,$\pm$\,0.03 & \textbf{0.90}\,$\pm$\,0.12 & \textbf{0.93}\,$\pm$\,0.08 \\
    \midrule
    \multirow{8}{*}{\shortstack[l]{P0\\Fridge}}
      & PPO            & 0.94\,$\pm$\,0.07 & 0.65\,$\pm$\,0.20 & \textbf{0.14}\,$\pm$\,0.09 & \textbf{0.14}\,$\pm$\,0.10 \\
      & PPO+Aug        & \textbf{0.93}\,$\pm$\,0.05 & \textbf{0.61}\,$\pm$\,0.24 & 0.19\,$\pm$\,0.06 & 0.19\,$\pm$\,0.06 \\
      & PPO+I-JEPA     & 0.95\,$\pm$\,0.02 & 0.82\,$\pm$\,0.08 & 0.59\,$\pm$\,0.15 & 0.60\,$\pm$\,0.14 \\
      & PPO+MAE        & 0.95\,$\pm$\,0.04 & 0.87\,$\pm$\,0.04 & 0.71\,$\pm$\,0.10 & 0.68\,$\pm$\,0.08 \\
      & PPO+DINOv2     & 0.96\,$\pm$\,0.03 & 0.93\,$\pm$\,0.02 & 0.80\,$\pm$\,0.05 & 0.77\,$\pm$\,0.08 \\
      & DreamerV3      & \textbf{0.99}\,$\pm$\,0.02 & \textbf{0.98}\,$\pm$\,0.04 & \textbf{0.97}\,$\pm$\,0.03 & \textbf{0.93}\,$\pm$\,0.08 \\
      & TD-MPC2        & \textbf{0.99}\,$\pm$\,0.02 & 0.85\,$\pm$\,0.08 & 0.61\,$\pm$\,0.13 & 0.63\,$\pm$\,0.12 \\
      & TD-MPC2+DINOv2 & 0.97\,$\pm$\,0.02 & 0.90\,$\pm$\,0.05 & 0.90\,$\pm$\,0.05 & 0.88\,$\pm$\,0.06 \\
    \midrule
    \multirow{8}{*}{\shortstack[l]{P3\\Bed}}
      & PPO            & 0.93\,$\pm$\,0.06 & 0.05\,$\pm$\,0.09 & \textbf{0.04}\,$\pm$\,0.05 & \textbf{0.04}\,$\pm$\,0.05 \\
      & PPO+Aug        & 0.94\,$\pm$\,0.08 & 0.23\,$\pm$\,0.21 & 0.19\,$\pm$\,0.18 & 0.19\,$\pm$\,0.18 \\
      & PPO+I-JEPA     & \textbf{0.78}\,$\pm$\,0.07 & 0.16\,$\pm$\,0.25 & 0.10\,$\pm$\,0.14 & 0.10\,$\pm$\,0.14 \\
      & PPO+MAE        & 0.86\,$\pm$\,0.07 & 0.25\,$\pm$\,0.22 & 0.14\,$\pm$\,0.20 & 0.14\,$\pm$\,0.20 \\
      & PPO+DINOv2     & 0.82\,$\pm$\,0.18 & 0.33\,$\pm$\,0.23 & 0.21\,$\pm$\,0.23 & 0.22\,$\pm$\,0.23 \\
      & DreamerV3      & 0.94\,$\pm$\,0.09 & \textbf{0.03}\,$\pm$\,0.03 & 0.13\,$\pm$\,0.12 & 0.14\,$\pm$\,0.07 \\
      & TD-MPC2        & 0.98\,$\pm$\,0.02 & 0.14\,$\pm$\,0.17 & 0.18\,$\pm$\,0.13 & 0.19\,$\pm$\,0.15 \\
      & TD-MPC2+DINOv2 & \textbf{1.00}\,$\pm$\,0.00 & \textbf{0.68}\,$\pm$\,0.16 & \textbf{0.46}\,$\pm$\,0.20 & \textbf{0.38}\,$\pm$\,0.18 \\
    \midrule
    \multirow{8}{*}{\shortstack[l]{P4\\Television}}
      & PPO            & 0.97\,$\pm$\,0.04 & 0.40\,$\pm$\,0.28 & 0.27\,$\pm$\,0.19 & 0.26\,$\pm$\,0.20 \\
      & PPO+Aug        & 0.92\,$\pm$\,0.07 & 0.41\,$\pm$\,0.36 & 0.26\,$\pm$\,0.24 & 0.26\,$\pm$\,0.25 \\
      & PPO+I-JEPA     & 0.90\,$\pm$\,0.06 & 0.50\,$\pm$\,0.28 & 0.38\,$\pm$\,0.23 & 0.34\,$\pm$\,0.23 \\
      & PPO+MAE        & 0.94\,$\pm$\,0.02 & \textbf{0.04}\,$\pm$\,0.07 & \textbf{0.05}\,$\pm$\,0.11 & \textbf{0.04}\,$\pm$\,0.09 \\
      & PPO+DINOv2     & 0.95\,$\pm$\,0.07 & 0.70\,$\pm$\,0.26 & 0.70\,$\pm$\,0.25 & 0.70\,$\pm$\,0.23 \\
      & DreamerV3      & \textbf{0.62}\,$\pm$\,0.43 & 0.74\,$\pm$\,0.31 & 0.79\,$\pm$\,0.19 & 0.77\,$\pm$\,0.23 \\
      & TD-MPC2        & \textbf{0.99}\,$\pm$\,0.02 & 0.70\,$\pm$\,0.28 & 0.35\,$\pm$\,0.23 & 0.34\,$\pm$\,0.31 \\
      & TD-MPC2+DINOv2 & 0.98\,$\pm$\,0.04 & \textbf{0.89}\,$\pm$\,0.04 & \textbf{0.86}\,$\pm$\,0.05 & \textbf{0.84}\,$\pm$\,0.06 \\
    \bottomrule
  \end{tabular}
\end{table}

\begin{table}[ht]
  \caption{Success rate in the full-shift house (L3 $=$ R4) in each of the
  three evaluation passes: the original ladder, the single-factor evaluation,
  and the reordered ladder; mean $\pm$ standard deviation over the five
  training seeds. Bold marks the highest and lowest value in each column.
  $^{\dagger n}$The two later passes use retrained models for $n$ of this agent
  type's 25 cells.}
  \label{tab:passes}
  \centering
  \small
  \resizebox{\textwidth}{!}{%
  \begin{tabular}{@{}lcccccc@{}}
    \toprule
    & \multicolumn{3}{c}{All five houses} & \multicolumn{3}{c}{P3} \\
    \cmidrule(lr){2-4} \cmidrule(l){5-7}
    Agent & Original & Single-change & Reordered & Original & Single-change & Reordered \\
    \midrule
    PPO            & 0.29\,$\pm$\,0.04 & 0.29\,$\pm$\,0.04 & 0.29\,$\pm$\,0.04 & \textbf{0.04}\,$\pm$\,0.05 & \textbf{0.04}\,$\pm$\,0.05 & \textbf{0.04}\,$\pm$\,0.05 \\
    PPO+Aug        & 0.34\,$\pm$\,0.08 & 0.34\,$\pm$\,0.08 & 0.34\,$\pm$\,0.08 & 0.19\,$\pm$\,0.18 & 0.19\,$\pm$\,0.18 & 0.19\,$\pm$\,0.18 \\
    PPO+I-JEPA     & 0.39\,$\pm$\,0.05 & 0.38\,$\pm$\,0.05 & 0.37\,$\pm$\,0.05 & 0.10\,$\pm$\,0.14 & 0.11\,$\pm$\,0.14 & 0.10\,$\pm$\,0.14 \\
    PPO+MAE        & \textbf{0.26}\,$\pm$\,0.06 & \textbf{0.25}\,$\pm$\,0.06 & \textbf{0.25}\,$\pm$\,0.06 & 0.14\,$\pm$\,0.20 & 0.14\,$\pm$\,0.20 & 0.14\,$\pm$\,0.20 \\
    PPO+DINOv2     & 0.66\,$\pm$\,0.07 & 0.66\,$\pm$\,0.08 & 0.66\,$\pm$\,0.07 & 0.22\,$\pm$\,0.23 & 0.22\,$\pm$\,0.23 & 0.22\,$\pm$\,0.24 \\
    DreamerV3$^{\dagger6}$ & 0.62\,$\pm$\,0.09 & 0.66\,$\pm$\,0.11 & 0.67\,$\pm$\,0.08 & 0.14\,$\pm$\,0.07 & 0.15\,$\pm$\,0.15 & 0.12\,$\pm$\,0.05 \\
    TD-MPC2$^{\dagger1}$   & 0.54\,$\pm$\,0.09 & 0.53\,$\pm$\,0.11 & 0.51\,$\pm$\,0.08 & 0.19\,$\pm$\,0.15 & 0.20\,$\pm$\,0.15 & 0.12\,$\pm$\,0.10 \\
    TD-MPC2+DINOv2 & \textbf{0.80}\,$\pm$\,0.06 & \textbf{0.81}\,$\pm$\,0.05 & \textbf{0.83}\,$\pm$\,0.06 & \textbf{0.38}\,$\pm$\,0.18 & \textbf{0.44}\,$\pm$\,0.17 & \textbf{0.50}\,$\pm$\,0.16 \\
    \bottomrule
  \end{tabular}
  }
\end{table}

\clearpage
\section{Single-factor results per agent and per house}
\label{app:single}

Table~\ref{tab:single-agent} gives each agent type's success rate after each
single change. Surface materials give the lowest success for every agent type
except PPO, for which the goal-object change gives a comparable value.

Table~\ref{tab:single-house} gives the same results for each house. Averaged
over agent types, the surface change takes success from 0.94 to 0.83 in P2,
0.94 to 0.62 in P1, 0.96 to 0.87 in P0, 0.90 to 0.32 in P3, and 0.94 to 0.63 in
P4. It gives the lowest single-factor success in four of the five houses; in
P0 the goal-object (0.85) and sky (0.86) changes give comparable values. With
any one house left out, the surface change still lowers success most on
average: by 21 to 33 points, against at most 10 for any other single change.
The effect is concentrated for the most robust agent type: TD-MPC2+DINOv2
changes by at most 0.04 in P2, P1, and P0, and falls from 1.00 to 0.70 in P3
and from 0.99 to 0.87 in P4. Each house has a single repaint, so the effect of
a repaint cannot be separated from that of its house.

\begin{table}[ht]
  \caption{Success rate after one factor changes, per agent type, over the runs
  that learned house A; mean $\pm$ standard deviation over the five training
  seeds. Bold marks the highest and lowest value in each column. ``Goal
  object'' covers the four houses with a swappable goal asset.}
  \label{tab:single-agent}
  \centering
  \small
  \resizebox{\textwidth}{!}{%
  \begin{tabular}{@{}lccccccc@{}}
    \toprule
    Agent & House A & Clutter & Lighting & Sky & Other objects & Goal object & Surfaces \\
    \midrule
    PPO            & 0.94\,$\pm$\,0.03 & 0.93\,$\pm$\,0.03 & 0.93\,$\pm$\,0.01 & \textbf{0.75}\,$\pm$\,0.08 & 0.86\,$\pm$\,0.04 & \textbf{0.58}\,$\pm$\,0.14 & 0.59\,$\pm$\,0.11 \\
    PPO+Aug        & 0.93\,$\pm$\,0.04 & 0.93\,$\pm$\,0.03 & 0.93\,$\pm$\,0.04 & 0.87\,$\pm$\,0.07 & 0.88\,$\pm$\,0.05 & 0.78\,$\pm$\,0.14 & 0.59\,$\pm$\,0.10 \\
    PPO+I-JEPA     & \textbf{0.89}\,$\pm$\,0.02 & \textbf{0.90}\,$\pm$\,0.02 & 0.90\,$\pm$\,0.02 & 0.90\,$\pm$\,0.03 & \textbf{0.84}\,$\pm$\,0.03 & 0.82\,$\pm$\,0.06 & 0.53\,$\pm$\,0.09 \\
    PPO+MAE        & 0.92\,$\pm$\,0.02 & 0.92\,$\pm$\,0.02 & \textbf{0.84}\,$\pm$\,0.05 & 0.86\,$\pm$\,0.03 & 0.88\,$\pm$\,0.03 & 0.87\,$\pm$\,0.04 & \textbf{0.47}\,$\pm$\,0.15 \\
    PPO+DINOv2     & 0.92\,$\pm$\,0.04 & 0.92\,$\pm$\,0.04 & 0.91\,$\pm$\,0.04 & 0.91\,$\pm$\,0.04 & 0.87\,$\pm$\,0.04 & 0.90\,$\pm$\,0.06 & 0.72\,$\pm$\,0.08 \\
    DreamerV3      & 0.91\,$\pm$\,0.05 & 0.92\,$\pm$\,0.02 & 0.90\,$\pm$\,0.03 & 0.91\,$\pm$\,0.04 & 0.89\,$\pm$\,0.06 & \textbf{0.98}\,$\pm$\,0.01 & 0.71\,$\pm$\,0.09 \\
    TD-MPC2        & \textbf{0.99}\,$\pm$\,0.01 & \textbf{0.99}\,$\pm$\,0.00 & 0.96\,$\pm$\,0.01 & 0.93\,$\pm$\,0.02 & \textbf{0.98}\,$\pm$\,0.01 & \textbf{0.98}\,$\pm$\,0.01 & 0.75\,$\pm$\,0.07 \\
    TD-MPC2+DINOv2 & \textbf{0.99}\,$\pm$\,0.01 & \textbf{0.99}\,$\pm$\,0.01 & \textbf{0.98}\,$\pm$\,0.01 & \textbf{0.98}\,$\pm$\,0.01 & 0.97\,$\pm$\,0.01 & \textbf{0.98}\,$\pm$\,0.01 & \textbf{0.89}\,$\pm$\,0.05 \\
    \bottomrule
  \end{tabular}
  }
\end{table}

\begin{table}[ht]
  \caption{Success rate after one factor changes, in each house, over the runs
  that learned house A; mean $\pm$ standard deviation over the training seeds.
  Bold marks the highest and lowest agent type in each column within each
  house; the last row of each house is the mean over the eight agent types. P2
  has no goal-only house.}
  \label{tab:single-house}
  \centering
  \scriptsize
  \setlength{\tabcolsep}{4pt}

\begin{tabular}{@{}llccccccc@{}}
\toprule
Pair & Agent & House A & Clutter & Lighting & Sky & Other objects & Goal object & Surfaces \\
\midrule
\multirow{9}{*}{\shortstack[l]{P2\\Bed}} & PPO & 0.91\,$\pm$\,0.03 & \textbf{0.91}\,$\pm$\,0.03 & 0.87\,$\pm$\,0.07 & 0.90\,$\pm$\,0.05 & \textbf{0.84}\,$\pm$\,0.05 & -- & 0.86\,$\pm$\,0.09 \\
 & PPO+Aug & 0.95\,$\pm$\,0.05 & 0.95\,$\pm$\,0.05 & 0.97\,$\pm$\,0.03 & 0.94\,$\pm$\,0.05 & 0.90\,$\pm$\,0.05 & -- & 0.85\,$\pm$\,0.07 \\
 & PPO+I-JEPA & 0.95\,$\pm$\,0.07 & 0.96\,$\pm$\,0.07 & 0.97\,$\pm$\,0.03 & 0.94\,$\pm$\,0.08 & 0.96\,$\pm$\,0.04 & -- & 0.86\,$\pm$\,0.12 \\
 & PPO+MAE & 0.92\,$\pm$\,0.12 & 0.92\,$\pm$\,0.12 & \textbf{0.72}\,$\pm$\,0.31 & \textbf{0.78}\,$\pm$\,0.15 & 0.85\,$\pm$\,0.15 & -- & \textbf{0.34}\,$\pm$\,0.35 \\
 & PPO+DINOv2 & \textbf{0.90}\,$\pm$\,0.04 & \textbf{0.91}\,$\pm$\,0.03 & 0.86\,$\pm$\,0.07 & 0.88\,$\pm$\,0.00 & \textbf{0.84}\,$\pm$\,0.04 & -- & 0.82\,$\pm$\,0.08 \\
 & DreamerV3 & \textbf{0.90}\,$\pm$\,0.20 & 0.93\,$\pm$\,0.14 & 0.90\,$\pm$\,0.20 & 0.95\,$\pm$\,0.10 & 0.92\,$\pm$\,0.16 & -- & 0.94\,$\pm$\,0.10 \\
 & TD-MPC2 & \textbf{1.00}\,$\pm$\,0.00 & \textbf{1.00}\,$\pm$\,0.00 & \textbf{1.00}\,$\pm$\,0.00 & 0.94\,$\pm$\,0.04 & \textbf{1.00}\,$\pm$\,0.00 & -- & 0.98\,$\pm$\,0.04 \\
 & TD-MPC2+DINOv2 & \textbf{1.00}\,$\pm$\,0.00 & \textbf{1.00}\,$\pm$\,0.00 & \textbf{1.00}\,$\pm$\,0.00 & \textbf{0.99}\,$\pm$\,0.02 & 0.99\,$\pm$\,0.02 & -- & \textbf{0.99}\,$\pm$\,0.02 \\
\cmidrule(l){2-9}
 & \textit{Mean of agent types} & 0.94\,$\pm$\,0.03 & 0.95\,$\pm$\,0.03 & 0.91\,$\pm$\,0.06 & 0.91\,$\pm$\,0.03 & 0.91\,$\pm$\,0.02 & -- & 0.83\,$\pm$\,0.04 \\
\midrule
\multirow{9}{*}{\shortstack[l]{P1\\Bed}} & PPO & 0.93\,$\pm$\,0.05 & 0.92\,$\pm$\,0.06 & 0.92\,$\pm$\,0.05 & 0.91\,$\pm$\,0.05 & 0.89\,$\pm$\,0.03 & \textbf{0.52}\,$\pm$\,0.22 & 0.79\,$\pm$\,0.09 \\
 & PPO+Aug & 0.94\,$\pm$\,0.06 & 0.94\,$\pm$\,0.05 & 0.94\,$\pm$\,0.05 & 0.91\,$\pm$\,0.06 & 0.92\,$\pm$\,0.06 & 0.83\,$\pm$\,0.16 & 0.49\,$\pm$\,0.26 \\
 & PPO+I-JEPA & 0.88\,$\pm$\,0.03 & 0.88\,$\pm$\,0.04 & 0.90\,$\pm$\,0.04 & 0.89\,$\pm$\,0.03 & 0.86\,$\pm$\,0.10 & 0.74\,$\pm$\,0.21 & \textbf{0.28}\,$\pm$\,0.07 \\
 & PPO+MAE & 0.93\,$\pm$\,0.02 & 0.94\,$\pm$\,0.02 & 0.86\,$\pm$\,0.05 & 0.90\,$\pm$\,0.02 & 0.92\,$\pm$\,0.03 & 0.99\,$\pm$\,0.02 & 0.52\,$\pm$\,0.22 \\
 & PPO+DINOv2 & 0.96\,$\pm$\,0.05 & 0.95\,$\pm$\,0.05 & 0.95\,$\pm$\,0.04 & 0.95\,$\pm$\,0.07 & 0.94\,$\pm$\,0.02 & 0.95\,$\pm$\,0.04 & 0.81\,$\pm$\,0.10 \\
 & DreamerV3 & \textbf{0.86}\,$\pm$\,0.21 & \textbf{0.84}\,$\pm$\,0.21 & \textbf{0.85}\,$\pm$\,0.16 & \textbf{0.81}\,$\pm$\,0.23 & \textbf{0.84}\,$\pm$\,0.18 & 0.98\,$\pm$\,0.04 & 0.44\,$\pm$\,0.18 \\
 & TD-MPC2 & \textbf{1.00}\,$\pm$\,0.00 & \textbf{1.00}\,$\pm$\,0.00 & \textbf{1.00}\,$\pm$\,0.00 & \textbf{1.00}\,$\pm$\,0.00 & \textbf{1.00}\,$\pm$\,0.00 & \textbf{1.00}\,$\pm$\,0.00 & 0.70\,$\pm$\,0.24 \\
 & TD-MPC2+DINOv2 & 0.99\,$\pm$\,0.02 & \textbf{1.00}\,$\pm$\,0.00 & 0.99\,$\pm$\,0.02 & 0.99\,$\pm$\,0.02 & 0.99\,$\pm$\,0.02 & 0.99\,$\pm$\,0.02 & \textbf{0.95}\,$\pm$\,0.04 \\
\cmidrule(l){2-9}
 & \textit{Mean of agent types} & 0.94\,$\pm$\,0.03 & 0.93\,$\pm$\,0.03 & 0.93\,$\pm$\,0.02 & 0.92\,$\pm$\,0.03 & 0.92\,$\pm$\,0.03 & 0.88\,$\pm$\,0.07 & 0.62\,$\pm$\,0.03 \\
\midrule
\multirow{9}{*}{\shortstack[l]{P0\\Fridge}} & PPO & 0.94\,$\pm$\,0.07 & \textbf{0.92}\,$\pm$\,0.07 & 0.94\,$\pm$\,0.07 & \textbf{0.50}\,$\pm$\,0.25 & 0.94\,$\pm$\,0.05 & \textbf{0.33}\,$\pm$\,0.21 & \textbf{0.56}\,$\pm$\,0.21 \\
 & PPO+Aug & \textbf{0.93}\,$\pm$\,0.05 & 0.93\,$\pm$\,0.05 & \textbf{0.92}\,$\pm$\,0.06 & 0.86\,$\pm$\,0.15 & 0.92\,$\pm$\,0.09 & 0.74\,$\pm$\,0.09 & 0.81\,$\pm$\,0.10 \\
 & PPO+I-JEPA & 0.94\,$\pm$\,0.02 & 0.94\,$\pm$\,0.04 & 0.95\,$\pm$\,0.02 & 0.90\,$\pm$\,0.04 & \textbf{0.81}\,$\pm$\,0.02 & 0.90\,$\pm$\,0.07 & 0.93\,$\pm$\,0.04 \\
 & PPO+MAE & 0.95\,$\pm$\,0.04 & 0.95\,$\pm$\,0.04 & 0.96\,$\pm$\,0.07 & 0.94\,$\pm$\,0.04 & 0.91\,$\pm$\,0.05 & 0.90\,$\pm$\,0.05 & 0.90\,$\pm$\,0.02 \\
 & PPO+DINOv2 & 0.96\,$\pm$\,0.03 & 0.96\,$\pm$\,0.03 & 0.97\,$\pm$\,0.02 & \textbf{0.96}\,$\pm$\,0.04 & 0.90\,$\pm$\,0.05 & 0.96\,$\pm$\,0.03 & 0.93\,$\pm$\,0.03 \\
 & DreamerV3 & \textbf{1.00}\,$\pm$\,0.00 & \textbf{1.00}\,$\pm$\,0.00 & \textbf{0.99}\,$\pm$\,0.02 & 0.92\,$\pm$\,0.09 & \textbf{1.00}\,$\pm$\,0.00 & \textbf{1.00}\,$\pm$\,0.00 & \textbf{1.00}\,$\pm$\,0.00 \\
 & TD-MPC2 & 0.99\,$\pm$\,0.02 & 0.99\,$\pm$\,0.02 & 0.98\,$\pm$\,0.02 & 0.86\,$\pm$\,0.12 & 0.98\,$\pm$\,0.02 & 0.99\,$\pm$\,0.02 & 0.90\,$\pm$\,0.10 \\
 & TD-MPC2+DINOv2 & 0.97\,$\pm$\,0.02 & 0.95\,$\pm$\,0.02 & 0.96\,$\pm$\,0.04 & 0.95\,$\pm$\,0.02 & 0.95\,$\pm$\,0.03 & 0.96\,$\pm$\,0.00 & 0.94\,$\pm$\,0.04 \\
\cmidrule(l){2-9}
 & \textit{Mean of agent types} & 0.96\,$\pm$\,0.01 & 0.95\,$\pm$\,0.02 & 0.96\,$\pm$\,0.01 & 0.86\,$\pm$\,0.06 & 0.92\,$\pm$\,0.01 & 0.85\,$\pm$\,0.03 & 0.87\,$\pm$\,0.03 \\
\midrule
\multirow{9}{*}{\shortstack[l]{P3\\Bed}} & PPO & 0.93\,$\pm$\,0.06 & 0.93\,$\pm$\,0.06 & 0.94\,$\pm$\,0.05 & \textbf{0.69}\,$\pm$\,0.22 & 0.87\,$\pm$\,0.08 & \textbf{0.50}\,$\pm$\,0.35 & \textbf{0.06}\,$\pm$\,0.08 \\
 & PPO+Aug & 0.94\,$\pm$\,0.08 & 0.93\,$\pm$\,0.08 & 0.90\,$\pm$\,0.08 & 0.72\,$\pm$\,0.18 & 0.86\,$\pm$\,0.06 & 0.62\,$\pm$\,0.39 & 0.38\,$\pm$\,0.23 \\
 & PPO+I-JEPA & \textbf{0.80}\,$\pm$\,0.08 & \textbf{0.79}\,$\pm$\,0.06 & \textbf{0.78}\,$\pm$\,0.07 & 0.87\,$\pm$\,0.08 & \textbf{0.72}\,$\pm$\,0.05 & 0.77\,$\pm$\,0.08 & 0.21\,$\pm$\,0.27 \\
 & PPO+MAE & 0.86\,$\pm$\,0.08 & 0.87\,$\pm$\,0.09 & 0.88\,$\pm$\,0.10 & 0.85\,$\pm$\,0.09 & 0.78\,$\pm$\,0.10 & 0.65\,$\pm$\,0.21 & 0.30\,$\pm$\,0.25 \\
 & PPO+DINOv2 & 0.82\,$\pm$\,0.18 & 0.82\,$\pm$\,0.18 & 0.82\,$\pm$\,0.18 & 0.80\,$\pm$\,0.19 & 0.77\,$\pm$\,0.17 & 0.72\,$\pm$\,0.17 & 0.35\,$\pm$\,0.21 \\
 & DreamerV3 & 0.89\,$\pm$\,0.14 & 0.92\,$\pm$\,0.07 & 0.85\,$\pm$\,0.18 & 0.89\,$\pm$\,0.13 & 0.83\,$\pm$\,0.21 & \textbf{0.99}\,$\pm$\,0.02 & 0.28\,$\pm$\,0.17 \\
 & TD-MPC2 & 0.97\,$\pm$\,0.03 & 0.95\,$\pm$\,0.02 & 0.91\,$\pm$\,0.08 & 0.93\,$\pm$\,0.03 & 0.95\,$\pm$\,0.04 & 0.94\,$\pm$\,0.05 & 0.32\,$\pm$\,0.25 \\
 & TD-MPC2+DINOv2 & \textbf{1.00}\,$\pm$\,0.00 & \textbf{0.99}\,$\pm$\,0.02 & \textbf{0.99}\,$\pm$\,0.02 & \textbf{0.99}\,$\pm$\,0.02 & \textbf{0.96}\,$\pm$\,0.03 & 0.98\,$\pm$\,0.04 & \textbf{0.70}\,$\pm$\,0.18 \\
\cmidrule(l){2-9}
 & \textit{Mean of agent types} & 0.90\,$\pm$\,0.05 & 0.90\,$\pm$\,0.05 & 0.89\,$\pm$\,0.05 & 0.84\,$\pm$\,0.07 & 0.84\,$\pm$\,0.06 & 0.77\,$\pm$\,0.10 & 0.32\,$\pm$\,0.07 \\
\midrule
\multirow{9}{*}{\shortstack[l]{P4\\Television}} & PPO & 0.97\,$\pm$\,0.04 & 0.96\,$\pm$\,0.04 & \textbf{0.98}\,$\pm$\,0.02 & \textbf{0.77}\,$\pm$\,0.20 & \textbf{0.75}\,$\pm$\,0.15 & 0.97\,$\pm$\,0.04 & 0.67\,$\pm$\,0.31 \\
 & PPO+Aug & 0.92\,$\pm$\,0.07 & 0.90\,$\pm$\,0.11 & 0.91\,$\pm$\,0.08 & 0.90\,$\pm$\,0.09 & 0.82\,$\pm$\,0.18 & 0.92\,$\pm$\,0.07 & 0.43\,$\pm$\,0.38 \\
 & PPO+I-JEPA & 0.90\,$\pm$\,0.05 & 0.91\,$\pm$\,0.05 & 0.88\,$\pm$\,0.04 & 0.90\,$\pm$\,0.05 & 0.87\,$\pm$\,0.06 & \textbf{0.88}\,$\pm$\,0.05 & 0.35\,$\pm$\,0.29 \\
 & PPO+MAE & 0.94\,$\pm$\,0.02 & 0.92\,$\pm$\,0.04 & \textbf{0.75}\,$\pm$\,0.10 & 0.83\,$\pm$\,0.08 & 0.92\,$\pm$\,0.05 & 0.94\,$\pm$\,0.04 & \textbf{0.30}\,$\pm$\,0.26 \\
 & PPO+DINOv2 & 0.97\,$\pm$\,0.03 & 0.95\,$\pm$\,0.02 & 0.97\,$\pm$\,0.02 & 0.95\,$\pm$\,0.03 & 0.92\,$\pm$\,0.04 & 0.96\,$\pm$\,0.03 & 0.70\,$\pm$\,0.25 \\
 & DreamerV3 & \textbf{0.85}\,$\pm$\,0.25 & \textbf{0.87}\,$\pm$\,0.20 & 0.91\,$\pm$\,0.16 & \textbf{0.99}\,$\pm$\,0.02 & 0.83\,$\pm$\,0.30 & 0.96\,$\pm$\,0.07 & \textbf{0.88}\,$\pm$\,0.17 \\
 & TD-MPC2 & 0.98\,$\pm$\,0.02 & \textbf{1.00}\,$\pm$\,0.00 & 0.93\,$\pm$\,0.03 & 0.94\,$\pm$\,0.04 & \textbf{0.96}\,$\pm$\,0.03 & \textbf{0.99}\,$\pm$\,0.02 & 0.86\,$\pm$\,0.08 \\
 & TD-MPC2+DINOv2 & \textbf{0.99}\,$\pm$\,0.02 & 0.99\,$\pm$\,0.02 & 0.96\,$\pm$\,0.03 & 0.98\,$\pm$\,0.02 & \textbf{0.96}\,$\pm$\,0.00 & 0.98\,$\pm$\,0.04 & 0.87\,$\pm$\,0.07 \\
\cmidrule(l){2-9}
 & \textit{Mean of agent types} & 0.94\,$\pm$\,0.03 & 0.94\,$\pm$\,0.03 & 0.91\,$\pm$\,0.02 & 0.91\,$\pm$\,0.01 & 0.88\,$\pm$\,0.03 & 0.95\,$\pm$\,0.01 & 0.63\,$\pm$\,0.06 \\
\bottomrule
\end{tabular}
\end{table}

\section{Goal-object control}
\label{app:goal}

Table~\ref{tab:goal} compares, for each agent type, success in L2noT, which
applies every L2 change except the goal asset, with success in L2, where the
goal asset changes too. Each agent is compared with itself in one evaluation
pass. In P2, whose goal cannot be swapped, L2noT and L2 are the same house;
there the difference is exactly zero for every run of the PPO-based agents and
at most 0.04 in magnitude, averaged over seeds, for the world models, whose
evaluation is stochastic.

\begin{table}[ht]
  \caption{Success rate with every L2 change except the goal asset (L2noT), and
  after the goal asset changes too (L2), over the 20 runs of each agent type in
  the four houses with a swappable goal; mean $\pm$ standard deviation over the
  five training seeds. Bold marks the highest and lowest value in each column.}
  \label{tab:goal}
  \centering
  \small
  \begin{tabular}{@{}lcc@{}}
    \toprule
    Agent & L2noT (goal unchanged) & L2 (goal changed) \\
    \midrule
    PPO            & 0.37\,$\pm$\,0.06 & \textbf{0.17}\,$\pm$\,0.05 \\
    PPO+Aug        & 0.38\,$\pm$\,0.10 & 0.23\,$\pm$\,0.10 \\
    PPO+I-JEPA     & 0.37\,$\pm$\,0.12 & 0.32\,$\pm$\,0.07 \\
    PPO+MAE        & \textbf{0.35}\,$\pm$\,0.05 & 0.27\,$\pm$\,0.04 \\
    PPO+DINOv2     & 0.66\,$\pm$\,0.09 & 0.63\,$\pm$\,0.07 \\
    DreamerV3      & 0.59\,$\pm$\,0.07 & 0.60\,$\pm$\,0.10 \\
    TD-MPC2        & 0.50\,$\pm$\,0.13 & 0.45\,$\pm$\,0.11 \\
    TD-MPC2+DINOv2 & \textbf{0.83}\,$\pm$\,0.06 & \textbf{0.78}\,$\pm$\,0.08 \\
    \bottomrule
  \end{tabular}
\end{table}

\section{Image measurements}
\label{app:image}

\begin{figure}[ht]
  \centering
  \includegraphics{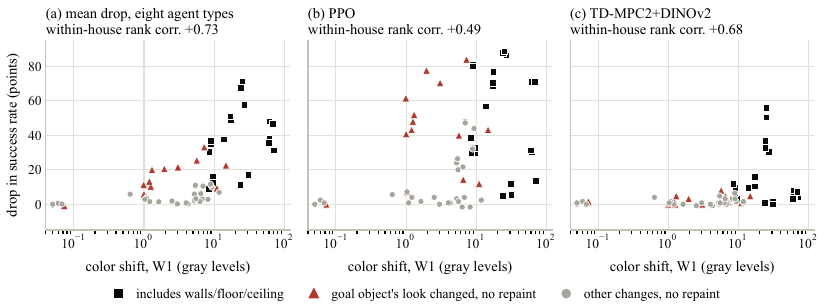}
  \caption{Color shift against the drop in success for every changed house (log
  scale; shifts below 0.05 are drawn at 0.05). Repainted houses (squares) carry
  most of the largest shifts and drops; replacing the goal asset (triangles)
  moves the colors little but can cost end-to-end PPO most of its success.}
  \label{fig:shift}
\end{figure}

Figure~\ref{fig:rgb} in Section~\ref{sec:imageshift} shows the pixel
distributions behind the color-shift measure for one house, and
Table~\ref{tab:shift} gives the measure for every house.
Table~\ref{tab:light-sky} separates the channels for lighting and sky and
compares the measure with the mean per-pixel difference, which can never be
smaller. Table~\ref{tab:corr} and Figure~\ref{fig:shift} relate it to the drop in
success over all 69 changed houses (five houses, 14 changes each, without the
missing goal-only house in P2), including the combinations used by the ladders.

Lighting and sky shift the color distribution by similar, small amounts, 5.6
and 5.4 gray levels on average across houses (lighting 3.0--8.1, sky 1.6--9.3;
Table~\ref{tab:shift}), yet their effects differ (Table~\ref{tab:single-agent}).
The new sky lowers PPO's success from 0.94 to 0.75 and lowers PPO+Aug,
PPO+MAE, and TD-MPC2 by 0.06--0.07; the new lighting leaves PPO almost
unchanged but lowers PPO+MAE from 0.92 to 0.84. The two changes differ in kind
(Table~\ref{tab:light-sky}). New lighting acts like a change of brightness: its
mean per-pixel difference equals its histogram distance in every house. The sky
also changes the light inside the room, but alters its color balance: in every
house its three channels shift less evenly than under new lighting (in P0, red
by 17.5 gray levels and blue by 2.9), and its per-pixel difference exceeds its
histogram distance, because some pixel values rise while others fall. PPO+Aug,
trained with hue and saturation jitter, keeps 0.87 success under the new sky,
against 0.75 for PPO.

\begin{table}[ht]
  \caption{Color shift (Wasserstein-1 distance from house A, mean over the red,
  green, and blue channels, in gray levels) for each single change and for the
  full shift. P2 has no goal-only house.}
  \label{tab:shift}
  \centering
  \small
  \begin{tabular}{@{}lccccc@{}}
    \toprule
    Change & P2 & P1 & P0 & P3 & P4 \\
    \midrule
    Clutter       & 0.0  & 4.2  & 0.1 & 0.0  & 0.1 \\
    Lighting      & 6.7  & 4.0  & 3.0 & 6.3  & 8.1 \\
    Sky           & 5.6  & 1.6  & 9.3 & 5.2  & 5.5 \\
    Other objects & 1.0  & 1.2  & 1.1 & 0.6  & 6.4 \\
    Goal object   & --   & 1.0  & 1.0 & 1.2  & 0.1 \\
    Surfaces      & 23.6 & 70.7 & 8.3 & 26.8 & 8.5 \\
    \midrule
    All changes (L3 = R4) & 30.7 & 68.3 & 9.0 & 25.0 & 17.0 \\
    \bottomrule
  \end{tabular}
\end{table}

\begin{table}[ht]
  \caption{Lighting and sky in detail: Wasserstein-1 distance from house A for
  each color channel and their mean (as in Table~\ref{tab:shift}), and the mean
  per-pixel difference between the same view in A and in the changed house, in
  gray levels. The two are equal exactly when no intensity level is crossed by
  pixels moving in opposite directions; this holds for lighting in every house
  and for the sky in none.}
  \label{tab:light-sky}
  \centering
  \small
  \begin{tabular}{@{}llccccc@{}}
    \toprule
    Change & Measure & P2 & P1 & P0 & P3 & P4 \\
    \midrule
    Lighting & W1, red               & 6.5 & 4.2 & 3.0 & 6.7 & 9.7 \\
      & W1, green             & 6.4 & 4.2 & 2.8 & 6.4 & 8.5 \\
      & W1, blue              & 7.3 & 3.6 & 3.1 & 5.7 & 6.2 \\
      & W1, mean of the three & 6.7 & 4.0 & 3.0 & 6.3 & 8.1 \\
      & Per-pixel difference  & 6.7 & 4.0 & 3.0 & 6.3 & 8.1 \\
    \midrule
    Sky      & W1, red               & 7.0 & 1.2 & 17.5 & 5.0 & 7.7 \\
      & W1, green             & 5.0 & 1.6 & 7.4  & 3.7 & 2.9 \\
      & W1, blue              & 4.7 & 2.1 & 2.9  & 6.9 & 5.7 \\
      & W1, mean of the three & 5.6 & 1.6 & 9.3  & 5.2 & 5.5 \\
      & Per-pixel difference  & 6.2 & 4.5 & 9.7  & 6.1 & 6.7 \\
    \bottomrule
  \end{tabular}
\end{table}

\begin{table}[ht]
  \caption{Within-house rank correlation between image measurements and the
  drop in success, averaged over houses. Permutation $p$ values are in
  parentheses. ``No repaint'' and ``repaint'' restrict the comparison to the
  indicated factors. The last row correlates the drop averaged over agent
  types.}
  \label{tab:corr}
  \centering
  \scriptsize
  \resizebox{\textwidth}{!}{%
  \begin{tabular}{@{}lccccc@{}}
    \toprule
    & \multicolumn{3}{c}{Color shift (W1)} & \multicolumn{2}{c}{Per-pixel difference} \\
    \cmidrule(lr){2-4} \cmidrule(l){5-6}
    Agent & all changes & no repaint & repaint & all changes & no repaint \\
    \midrule
    PPO            & +0.49 ($<$0.001) & +0.10 (0.522) & +0.00 (0.990) & +0.67 ($<$0.001) & +0.40 (0.012) \\
    PPO+Aug        & +0.67 ($<$0.001) & +0.23 (0.160) & +0.16 (0.490) & +0.82 ($<$0.001) & +0.47 (0.003) \\
    PPO+I-JEPA     & +0.61 ($<$0.001) & +0.08 (0.612) & +0.21 (0.352) & +0.75 ($<$0.001) & +0.30 (0.059) \\
    PPO+MAE        & +0.76 ($<$0.001) & +0.50 (0.001) & +0.11 (0.634) & +0.85 ($<$0.001) & +0.57 ($<$0.001) \\
    PPO+DINOv2     & +0.65 ($<$0.001) & +0.17 (0.296) & +0.13 (0.583) & +0.76 ($<$0.001) & +0.24 (0.133) \\
    DreamerV3      & +0.60 ($<$0.001) & +0.27 (0.097) & +0.13 (0.559) & +0.60 ($<$0.001) & +0.28 (0.082) \\
    TD-MPC2        & +0.79 ($<$0.001) & +0.70 ($<$0.001) & $-$0.16 (0.480) & +0.85 ($<$0.001) & +0.71 ($<$0.001) \\
    TD-MPC2+DINOv2 & +0.68 ($<$0.001) & +0.24 (0.142) & +0.08 (0.720) & +0.73 ($<$0.001) & +0.37 (0.019) \\
    \midrule
    All agents (average drop) & +0.73 ($<$0.001) & +0.37 (0.021) & +0.11 (0.627) & +0.86 ($<$0.001) & +0.62 ($<$0.001) \\
    \bottomrule
  \end{tabular}
  }
\end{table}

\clearpage
\section{Training budget}
\label{app:budget}

At 150k steps PPO+Aug had not converged in house A (0.84, against 0.92 for
PPO), and 9 of DreamerV3's 25 runs had not learned house A (5 at 300k steps). At
300k steps PPO+Aug matches PPO in house A (0.93), and its advantage over PPO in
success at L2 shrinks from 20 to 6 points (Table~\ref{tab:budget}).

\begin{table}[ht]
  \caption{Success rate in house A and at L2 after 150k and 300k training
  steps, every run; mean $\pm$ standard deviation over the five training seeds.
  Bold marks the highest and lowest value in each column.}
  \label{tab:budget}
  \centering
  \small
  \begin{tabular}{@{}lcccc@{}}
    \toprule
    & \multicolumn{2}{c}{150k steps} & \multicolumn{2}{c}{300k steps} \\
    \cmidrule(lr){2-3} \cmidrule(l){4-5}
    Agent & A & L2 & A & L2 \\
    \midrule
    PPO       & 0.92\,$\pm$\,0.03 & \textbf{0.27}\,$\pm$\,0.04 & 0.93\,$\pm$\,0.02 & \textbf{0.29}\,$\pm$\,0.04 \\
    PPO+Aug   & 0.84\,$\pm$\,0.05 & 0.47\,$\pm$\,0.07 & 0.93\,$\pm$\,0.04 & 0.35\,$\pm$\,0.07 \\
    DreamerV3 & \textbf{0.71}\,$\pm$\,0.13 & \textbf{0.63}\,$\pm$\,0.11 & \textbf{0.76}\,$\pm$\,0.10 & \textbf{0.60}\,$\pm$\,0.10 \\
    TD-MPC2   & \textbf{0.99}\,$\pm$\,0.01 & 0.55\,$\pm$\,0.09 & \textbf{0.99}\,$\pm$\,0.00 & 0.54\,$\pm$\,0.09 \\
    \bottomrule
  \end{tabular}
\end{table}

\section{Declared comparisons}
\label{app:declared}

The declared tests compare retained success, defined as success in a ladder
condition divided by success in house A, with a permutation test that shuffles
agent-type labels within each house, over the runs that learned house A. Holm
correction~\cite{holm1979} is applied within declared families. The
original-four family was fixed on 2026-09-05 before pretrained-encoder agents
were trained. The DINOv2 family was fixed after the means for PPO+DINOv2 were
observed. The TD-MPC2+DINOv2 family was fixed after that agent's means and
confidence intervals were observed. A positive value means the second agent
retains more of its training-house success. Table~\ref{tab:declared} lists
every family and its comparisons.

\begin{table}[ht]
  \caption{Declared comparisons on the original ladder at 300k steps.
  Difference is the mean difference in retained success, in percentage points.
  Bold values remain significant after Holm correction within the stated
  family.}
  \label{tab:declared}
  \centering
  \small
  \begin{tabular}{@{}llccc@{}}
    \toprule
    Agent 1 & Agent 2 & Difference (points) & $p$ & $p_{\mathrm{Holm}}$ \\
    \midrule
    \multicolumn{5}{@{}l}{\textit{All pairs, retained success at L2 (28 comparisons)}} \\
    PPO        & PPO+Aug        & +6.7  & 0.107 & 0.640 \\
    PPO        & PPO+I-JEPA     & +10.9 & 0.040 & 0.319 \\
    PPO        & PPO+MAE        & $-$3.4  & 0.642 & 1.000 \\
    PPO        & PPO+DINOv2     & +40.4 & $<$0.0001 & \textbf{0.0003} \\
    PPO        & DreamerV3      & +33.5 & 0.0001 & \textbf{0.002} \\
    PPO        & TD-MPC2        & +22.4 & $<$0.0001 & \textbf{0.0008} \\
    PPO        & TD-MPC2+DINOv2 & +52.1 & $<$0.0001 & \textbf{0.0003} \\
    PPO+Aug    & PPO+I-JEPA     & +4.2  & 0.469 & 1.000 \\
    PPO+Aug    & PPO+MAE        & $-$10.1 & 0.173 & 0.866 \\
    PPO+Aug    & PPO+DINOv2     & +33.7 & $<$0.0001 & \textbf{0.0003} \\
    PPO+Aug    & DreamerV3      & +26.8 & 0.003 & \textbf{0.038} \\
    PPO+Aug    & TD-MPC2        & +15.7 & 0.006 & 0.068 \\
    PPO+Aug    & TD-MPC2+DINOv2 & +45.4 & $<$0.0001 & \textbf{0.0003} \\
    PPO+I-JEPA & PPO+MAE        & $-$14.3 & 0.020 & 0.194 \\
    PPO+I-JEPA & PPO+DINOv2     & +29.5 & 0.0001 & \textbf{0.002} \\
    PPO+I-JEPA & DreamerV3      & +22.6 & 0.003 & \textbf{0.034} \\
    PPO+I-JEPA & TD-MPC2        & +11.5 & 0.048 & 0.337 \\
    PPO+I-JEPA & TD-MPC2+DINOv2 & +41.2 & $<$0.0001 & \textbf{0.0003} \\
    PPO+MAE    & PPO+DINOv2     & +43.8 & $<$0.0001 & \textbf{0.0003} \\
    PPO+MAE    & DreamerV3      & +36.8 & $<$0.0001 & \textbf{0.0008} \\
    PPO+MAE    & TD-MPC2        & +25.7 & 0.0002 & \textbf{0.004} \\
    PPO+MAE    & TD-MPC2+DINOv2 & +55.5 & $<$0.0001 & \textbf{0.0003} \\
    PPO+DINOv2 & DreamerV3      & $-$6.9  & 0.373 & 1.000 \\
    PPO+DINOv2 & TD-MPC2        & $-$18.1 & 0.005 & 0.060 \\
    PPO+DINOv2 & TD-MPC2+DINOv2 & +11.7 & 0.019 & 0.194 \\
    DreamerV3  & TD-MPC2        & $-$11.1 & 0.174 & 0.866 \\
    DreamerV3  & TD-MPC2+DINOv2 & +18.7 & 0.002 & \textbf{0.024} \\
    TD-MPC2    & TD-MPC2+DINOv2 & +29.8 & $<$0.0001 & \textbf{0.0003} \\
    \midrule
    \multicolumn{5}{@{}l}{\textit{Original four agents, retained success at L2 (6 comparisons)}} \\
    PPO       & PPO+Aug   & +6.7  & 0.107 & 0.213 \\
    PPO       & DreamerV3 & +33.5 & 0.0001 & \textbf{0.0004} \\
    PPO       & TD-MPC2   & +22.4 & $<$0.0001 & \textbf{0.0002} \\
    PPO+Aug   & DreamerV3 & +26.8 & 0.003 & \textbf{0.012} \\
    PPO+Aug   & TD-MPC2   & +15.7 & 0.006 & \textbf{0.018} \\
    DreamerV3 & TD-MPC2   & $-$11.1 & 0.174 & 0.213 \\
    \midrule
    \multicolumn{5}{@{}l}{\textit{PPO+DINOv2 versus the six earlier agents, retained success at L1 (6 comparisons)}} \\
    PPO        & PPO+DINOv2 & +22.7 & 0.0002 & \textbf{0.001} \\
    PPO+Aug    & PPO+DINOv2 & +22.2 & 0.003 & \textbf{0.009} \\
    PPO+I-JEPA & PPO+DINOv2 & +27.6 & 0.0002 & \textbf{0.0008} \\
    PPO+MAE    & PPO+DINOv2 & +41.1 & $<$0.0001 & \textbf{0.0001} \\
    DreamerV3  & PPO+DINOv2 & +14.3 & 0.045 & \textbf{0.045} \\
    TD-MPC2    & PPO+DINOv2 & +13.4 & 0.021 & \textbf{0.043} \\
    \midrule
    \multicolumn{5}{@{}l}{\textit{TD-MPC2+DINOv2 versus each agent, retained success at L1 (7 comparisons)}} \\
    PPO        & TD-MPC2+DINOv2 & +34.0 & $<$0.0001 & \textbf{0.0001} \\
    PPO+Aug    & TD-MPC2+DINOv2 & +33.6 & $<$0.0001 & \textbf{0.0001} \\
    PPO+I-JEPA & TD-MPC2+DINOv2 & +38.9 & $<$0.0001 & \textbf{0.0001} \\
    PPO+MAE    & TD-MPC2+DINOv2 & +52.4 & $<$0.0001 & \textbf{0.0001} \\
    PPO+DINOv2 & TD-MPC2+DINOv2 & +11.3 & 0.009 & \textbf{0.009} \\
    DreamerV3  & TD-MPC2+DINOv2 & +25.6 & 0.0001 & \textbf{0.0002} \\
    TD-MPC2    & TD-MPC2+DINOv2 & +24.7 & $<$0.0001 & \textbf{0.0001} \\
    \bottomrule
  \end{tabular}
\end{table}

\end{document}